%% file: iclr2027_conference.tex
\documentclass{article} % For LaTeX2e
\usepackage{iclr2027_conference,times}
\usepackage{booktabs} 
\usepackage{float}
\usepackage{multirow}
\usepackage{graphicx}
\usepackage{amssymb}
\usepackage{tikz}
\usepackage{array}
\input{figures/styles}
\input{math_commands.tex}

\usepackage{hyperref}
\usepackage{url}
\usepackage{tcolorbox,listings,graphicx}
\usepackage{placeins}

\usepackage[table]{xcolor}
\definecolor{winQwen}{HTML}{D3E1F2}   
\definecolor{winMini}{HTML}{D7E9D8}  
\definecolor{winLFM} {HTML}{F7E7CF}

\title{Small Agents with Semantic Search: Efficient Multilingual Code Localization}
\author{Maxence Lasbordes$^{1,2}$ \quad
Aarush Sinha$^{1,3}$ \quad
Raphael Sourty$^{1}$ \quad
Amélie Chatelain$^{1}$
\AND
Djamé Seddah$^{2}$ \\[12pt]
$^{1}$LightOn \\
$^{2}$Inria Paris\\
$^{3}$University of Copenhagen
}

\usepackage{xspace} % added by Djamé

\newcommand{\grep}{\texttt{grep}\xspace}

\newcommand{\colgrep}{\texttt{colgrep}\xspace}

\iclrfinalcopy % Uncomment for camera-ready version, but NOT for submission.

\begin{document}
\maketitle

\begin{abstract}

Locating relevant files from natural-language requests is a core subtask for agents operating over code repositories. We investigate whether this task can be delegated to compact, specialized models to enable on-device search while reducing the token usage, latency, and inference cost of larger agents. We show that semantic search improves file localization, with gains in accuracy, cross-language transfer, and inference efficiency. To study this setting, we introduce a training framework for file-localization agents built around \colgrep, a local semantic search tool based on late-interaction retrieval models. Our recipe combines weighted supervised fine-tuning on teacher trajectories, assigning turn-level credit based on retrieval outcomes, with reinforcement learning on localization quality. We train three model families with fewer than two billion parameters to formulate search queries, inspect retrieved content, and identify relevant files. On localization tasks derived from SWE-bench Lite and Multi-SWE-bench Flash, \colgrep-equipped agents substantially improve over their base models and outperform corresponding \grep-based agents. In addition to improving localization accuracy, \colgrep reduces mean end-to-end trajectory latency by 44.1\% on CPU while using 29.1\% fewer tokens, and enables better generalization to programming languages unseen during fine-tuning. These results suggest that compact, tool-specialized localization agents can provide an efficient interface between natural-language requests and large codebases. 

% We will release model checkpoints, software, data, and training pipelines upon acceptance.

\end{abstract}

\section{Introduction}
\label{sec:introduction}

Coding agents must first locate where to make a change in a repository before producing the change itself. Finding those files may require repeated directory listings, lexical searches, and file reads, making localization a substantial part of the overall repair process. CodeGrep \citep{chen2026codegrep}, for example, reports 23 interaction rounds and 631K tokens per resolved SWE-bench Verified issue for a 30B OpenHands agent \citep{jimenez2024swebench}. A natural way to make this process more efficient is to use a small, specialized agent dedicated to the localization step. This agent can return candidate files, allowing a larger and more expensive model to begin from a focused context rather than searching the repository from scratch. Our paper isolates file localization and measures localization accuracy, trajectory length, and latency.

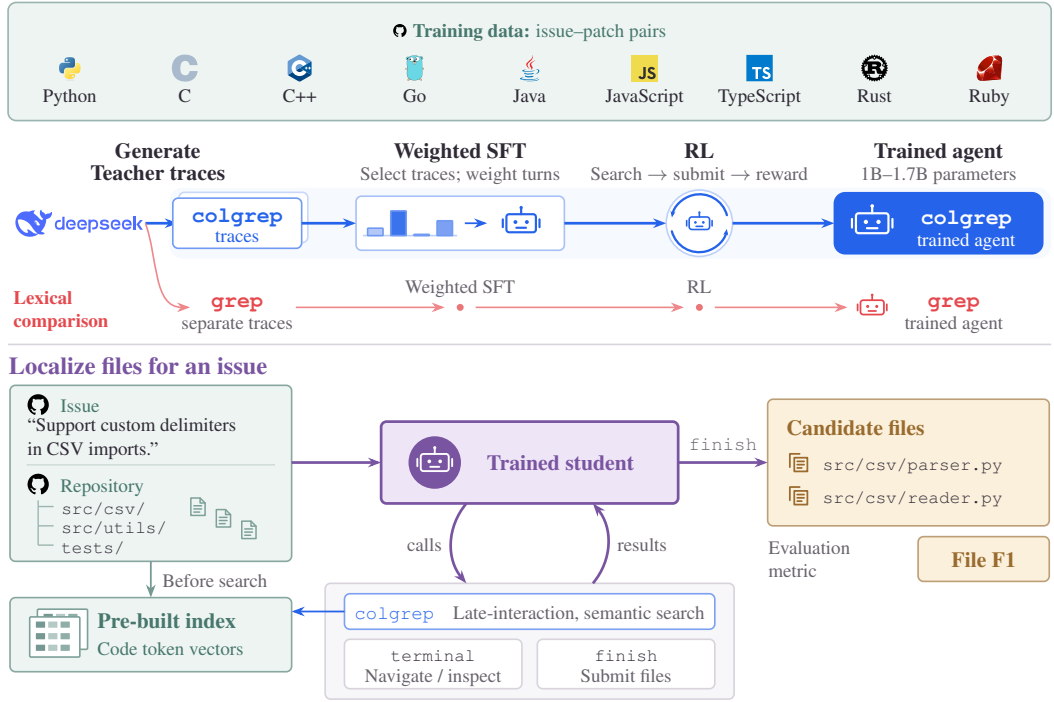
\begin{figure*}[!tbp]
    \centering
    \resizebox{\textwidth}{!}{%
        \input{figures/overview_codeonly}
    }
    \caption{\textbf{Training compact agents for file localization.} Given an issue and repository, the policy searches indexed code with \colgrep, inspects candidates in a read-only terminal, and submits file paths with \texttt{finish}. The \grep branch represents separately generated demonstrations and lexical-only policies.}
    \label{fig:shared-loop}
    \vspace{-1em}
\end{figure*}

This separation is also consistent with prior work showing that repository-level repair can be decomposed into code localization and editing \citep{xia2024agentless,xie2025swe}. Treating localization as a standalone learned decision problem, in turn, lets us compare two interfaces available to compact agents: lexical tools \citep{sutawika2026codescout} and semantic retrievers \citep{vijay2025orion,reddy2025swerankplus}. These interfaces place different demands on the policy. A lexical tool such as \grep requires a string or regular expression, whereas semantic search accepts a natural-language description. The latter may reduce the need to infer repository-specific identifiers \citep{mahmud2024gui}. However, lexical retrieval remains effective with capable policies and suitable access to results \citep{hsu2026rethinkingagenticsearchpiserini}, and task-specific training may further narrow the difference. We therefore ask: \emph{How do search interfaces and training signals shape file-localization policies below two billion parameters?}

To study this question, we introduce a two-stage training framework for compact localization policies centered on \colgrep, a local search tool based on late-interaction retrieval \citep{next-plaid,khattab2020colbert}. A \colgrep call combines a natural-language query with an optional regular-expression filter, allowing the policy to pair a behavioral description with an exact identifier. Supervised fine-tuning (SFT) teaches the policy to formulate queries, inspect retrieved code, and produce file submissions, with assistant turns weighted by their retrieval outcomes. Reinforcement learning (RL) then directly optimizes the quality of the final submission through repository interaction. We apply this framework to three model families with 1--1.7B parameters and evaluate file localization on SWE-bench Lite and Multi-SWE-bench Flash \citep{jimenez2024swebench,zan2026multiswebench}.

Our contributions are: \textbf{(i)} a systematic comparison of semantic and lexical search for compact file-localization agents, demonstrating consistent gains from semantic search across training stages on SWE-bench Lite and Multi-SWE-bench Flash; \textbf{(ii)} a cross-language generalization study showing that semantic search enables substantially stronger transfer than lexical search to seven programming languages unseen during fine-tuning; \textbf{(iii)} an efficiency analysis demonstrating that, after training, semantic search improves CPU/GPU latency and token usage alongside localization accuracy ; and \textbf{(iv)} a training recipe for small, efficient semantic-search agents, including a weighted SFT objective that improves file F1 across all three student models and both benchmarks at unchanged fine-tuning compute.

\section{Related Work}

\paragraph{Code Localization.} Separating localization from editing is well established in repository-level issue resolution. Agentless decomposes the workflow into localization, repair, and validation \citep{xia2024agentless}, while SWE-Fixer trains separate file-retrieval and editing modules \citep{xie2025swe}. Beyond this decomposition, interactive methods expose additional repository structure or support repeated retrieval. Building on graph-based localization, LocAgent uses graph-guided navigation \citep{chen2025locagent}, while OrcaLoca combines code-graph search with prioritized actions \citep{yu2025orcaloca}. More recently, SweRank+ complements these approaches with multilingual ranking and multi-turn localization \citep{reddy2025swerankplus}. In parallel, learning dedicated search policies has emerged as an active direction: ToolTrain combines rejection-sampled SFT with tool-integrated RL \citep{ma2025tool}; CodeScout \citep{sutawika2026codescout} trains code-search agents using a standard Unix terminal; and CodeGrep \citep{chen2026codegrep} learns multi-turn lexical retrieval for a downstream coding agent. Together, these lines of work motivate our focus on the interaction between compact policies, semantic access, and supervision.

\paragraph{Search Interfaces and Agent Training.} Agent performance depends on the tools and observations exposed by the harness, as studied by SWE-agent \citep{yang2024swe}. Lexical tools provide exact matching and familiar navigation, while semantic tools enable retrieval from descriptions that need not share the target's identifiers. \colgrep implements such semantic retrieval using late interaction in the ColBERT family \citep{khattab2020colbert}, with an indexing backend based on PLAID \citep{santhanam2022plaid}. We evaluate semantic-equipped policies against terminal-only policies, treating the retriever and its tool feedback as part of the intervention. For training, teacher distillation supplies executable examples of search and submission. Whole-trajectory filtering determines which demonstrations are retained, whereas weighted supervision adjusts their contribution to the loss. Importance-weighted SFT provides related motivation \citep{qin2025supervised}, but our weights are fixed task-specific credits for discovery and inspection rather than model-derived importance ratios. Subsequent RL uses GRPO \citep{shao2024deepseekmath} with dynamic removal of uninformative rollout groups, following the sampling principle used in DAPO \citep{yu2026dapo}. We study the resulting policies and reward trade-offs rather than proposing a new policy-optimization algorithm.

\section{Task and Experimental Setup}

\subsection{Agent Harness and Search Interfaces}
\label{sec:harness}

We design our own harness for the code localization task, as presented in Figure \ref{fig:shared-loop}. The harness specifies the system prompt, tool schemas (Appendix \ref{sec:harness_prompts}), read-only sandbox, sampling parameters, and interaction budget, thereby defining the actions available to the model and the feedback it receives. Following prior work favoring small tool sets for localization \citep{sutawika2026codescout,zhang2026toolenoughreinforcementlearning}, we keep the harness minimal. We use the same localization harness for both training and evaluation.

\paragraph{Colgrep: Semantic Search.}
\colgrep \citep{next-plaid} is the semantic search engine used in our harness. It is a single Rust binary built on NextPlaid, a multi-vector search engine implementing the PLAID index, and runs entirely locally without a server or API. We use \colgrep 1.7.0 and build all indexes once before training and evaluation, so inference-time retrieval only requires query encoding and index lookup. We build one index per repository at the instance's base commit. \colgrep parses source files with tree-sitter and splits them into code units---functions, methods, classes, constants, and remaining blocks---across 35 languages. In all experiments, we use the same \texttt{LateOn-Code-edge} encoder \citep{LateOn-Code}, a 17M-parameter late-interaction model \citep{khattab2020colbert}. Each code unit is represented as a sequence of token vectors and scored with MaxSim: for each query token, MaxSim selects the highest similarity to any token in the candidate unit and sums these scores across the query. This preserves lexical, identifier, and semantic matches while enabling efficient CPU indexing and search.

\paragraph{Tools and session protocol.}
The agent has access to three tools. \colgrep, as detailed above, performs semantic search over a pre-built index from a natural-language query, with optional features such as \emph{path}, file-type \emph{include}, or regex \emph{pattern}. \texttt{terminal} provides a read-only shell with standard Unix utilities for navigation, lexical search, and file inspection. Access is restricted to the repository at the instance's base commit. \texttt{finish} submits file paths with optional line ranges or symbol names. The system prompt asks the model to identify locations relevant to an issue without modifying the repository; the user turn provides the issue description and repository root listing. The model may interact for up to 10 turns, with at most 5 parallel tool calls per turn. Malformed tool calls receive corrective feedback and may be retried up to 3 times. A reminder is injected near the end of the interaction budget, and if \texttt{finish} has not yet been called, a rescue prompt requests a final submission. If the model still fails to submit, the inspected files are used as a fallback submission so that every session produces a scorable output.

\paragraph{Retrieval interface.} A search typically uses a \texttt{query} argument containing the natural-language query passed to the semantic search model and returns the top-$k$ matching units. The optional \texttt{k} argument controls the number of results, with a default of 10 and a maximum of 25. Results are returned as \texttt{file:start-end} locations with short snippets. Three optional arguments can further restrict the candidate set before ranking: \texttt{path} limits retrieval to a subdirectory or file, \texttt{include} applies a file glob, and \texttt{pattern} applies a \grep-style regular-expression pre-filter. This supports hybrid lexical and semantic search: the regex selects candidates exactly, while the semantic model ranks them by the intent expressed in \texttt{query}. The policy can therefore combine an identifier, symbol, or error string in \texttt{pattern} with a natural-language description of the desired behavior in \texttt{query}.

\begin{table*}[!tbp]
\centering\small
\setlength{\tabcolsep}{3pt}
\renewcommand{\arraystretch}{1.05}

\newcolumntype{M}{>{\centering\arraybackslash}p{0.75cm}}
\newcolumntype{B}{>{\columncolor{cyan!10}\centering\arraybackslash}p{0.75cm}}

\resizebox{\textwidth}{!}{%
\begin{tabular}{lll B M M M M B M M}
\toprule
& & &
\multicolumn{5}{c}{\textbf{SWE-bench Lite}} &
\multicolumn{3}{c}{\textbf{Multi-SWE-bench Flash}} \\
\cmidrule(lr){4-8}
\cmidrule(lr){9-11}

& & &
\multicolumn{3}{c}{\textbf{File}} &
\multicolumn{1}{c}{\textbf{Class}} &
\multicolumn{1}{c}{\textbf{Function}} &
\multicolumn{3}{c}{\textbf{File}} \\
\cmidrule(lr){4-6}
\cmidrule(lr){7-7}
\cmidrule(lr){8-8}
\cmidrule(lr){9-11}

\multicolumn{1}{c}{\textbf{Model}} &
\multicolumn{1}{c}{\textbf{Training}} &
\multicolumn{1}{c}{\textbf{Search tool}} &
\cellcolor{cyan!10}\textbf{F1} & R & P & F1 & F1 &
\cellcolor{cyan!10}\textbf{F1} & R & P \\
\midrule

\multirow{6}{*}{Qwen3-1.7B}
& \multirow{2}{*}{Base}
& \grep
& 18.75 & 26.13 & 16.66 & 12.21 & \textbf{5.65}
& 12.38 & 14.32 & 14.31 \\
& & \colgrep
& \textbf{34.51} & \textbf{56.53} & \textbf{29.30}
& \textbf{13.64} & 2.81
& \textbf{29.39} & \textbf{43.18} & \textbf{31.45} \\

\cmidrule(lr){2-11}

& \multirow{2}{*}{Weighted SFT}
& \grep
& 50.13 & 60.93 & 45.60 & \textbf{33.58} & \textbf{18.00}
& 32.84 & 36.47 & 38.53 \\
& & \colgrep
& \textbf{55.09} & \textbf{71.67} & \textbf{48.46}
& 28.62 & 8.00
& \textbf{44.58} & \textbf{54.26} & \textbf{47.31} \\

\cmidrule(lr){2-11}

& \multirow{2}{*}{+ RL: file F1}
& \grep
& 63.75 & 70.60 & 60.64 & \textbf{45.05} & \textbf{25.05}
& 50.16 & 49.14 & 62.84 \\
& & \colgrep
& \textbf{67.29} & \textbf{72.67} & \textbf{64.77}
& 33.09 & 9.53
& \textbf{54.00} & \textbf{52.81} & \textbf{67.07} \\
\midrule
DeepSeek-V4-Flash-0731 & - &
  \colgrep & 63.96 & 90.73 & 53.26 & 50.30 & 30.51
  & 59.12 & 74.85 & 57.34 \\
Qwen3-32B & - & \colgrep
& 63.50 & 80.33 & 57.00 & 36.59 & 14.15
& 54.02 & 60.48 & 60.43 \\

\bottomrule
\end{tabular}
}
\caption{Comparison of \grep{} and \colgrep{} across training stages for Qwen3-1.7B. Bold indicates the better search interface for each metric within a training stage.}
\label{tab:qwen_grep_vs_colgrep_training}
\vspace{-1em}
\end{table*}

\subsection{Training}
\label{sec:training}

We construct a corpus spanning multiple programming languages by pairing code repositories with issue descriptions and corresponding gold commits. For SFT, we sample issue--patch instances from SWE-Fixer \citep{xie2025swe} and SWE-rebench \citep{badertdinov2025swerebenchautomatedpipelinetask}. The resulting corpus spans 2,688 repositories and 13,452 distinct repository--commit checkouts, comprising 4,288 Python instances and 9,164 instances across seven additional languages: C, C++, Go, Java, JavaScript, Rust, and TypeScript, as shown in Table~\ref{tab:sft_languages}. For RL, we construct a separate corpus by mining bug-fix commits from public code repositories, yielding approximately 11,000 instances from 33 repositories across Python, Go, JavaScript, TypeScript, C, Ruby, Rust, and Java. For each instance, the parent commit defines the repository state used as the search environment, the issue or commit description provides the query, and the corresponding patch defines the target locations. We deduplicate by instance ID and repository/base-commit/patch hashes, decontaminate by excluding benchmark repositories and shared commit histories, and screen problem statements with long-$n$-gram matching. We apply our recipe to three students, ranging from 1--1.7B parameters: Qwen3-1.7B \citep{qwen3technicalreport}, MiniCPM5-1B \citep{minicpm4}, and LFM2.5-1.2B-Instruct \citep{liquidai2025lfm2}.

\subsection{Evaluation Protocol}
\label{sec:eval}

We evaluate file localization from code problem statements by adapting benchmarks originally designed for end-to-end issue resolution, SWE-bench Lite \citep{jimenez2024swebench} and Multi-SWE-bench Flash \citep{zan2026multiswebench}. Each instance pairs a GitHub issue with the human-written pull request that resolves it at a fixed base commit. We use the eligible pre-patch files and line ranges modified by that pull request as gold localization targets and ask the agent to identify them rather than produce a fix. SWE-bench Lite contains 300 issues from 12 Python repositories. Multi-SWE-bench Flash contains 300 instances spanning 24 repositories and seven languages: C, C++, Java, Go, Rust, JavaScript, and TypeScript. Our primary metric is file-level F1, reflecting our goal of selecting relevant files for downstream inspection and editing while balancing coverage and precision. For each instance, we compute file-level precision, recall, and F1, with empty predictions scored as zero, and macro-average scores across instances. On SWE-bench Lite, we additionally report class- and function-level F1, defining gold classes and functions as the tree-sitter definitions enclosing modified lines at the base commit.

We evaluate our students using a 10-turn search budget, native function calling through vLLM \citep{kwon2023efficient}, a 32k-token context, and at most 2,048 completion tokens per turn. Sampling uses temperature 0.6 and top-$p=0.95$, with \texttt{thinking} disabled where supported. We compute the average of five seeds for each evaluation.

\section{Training Compact Localization Agents}
\label{sec:training-sft-rl}

\subsection{Supervised Fine-Tuning from Teacher Search Trajectories}
\label{sec:sft}

We use SFT to teach our student models to search, inspect results, and submit relevant locations through the harness. Tool descriptions alone are often insufficient: models may continue to rely on familiar shell commands such as \grep even when semantic search is available. We therefore use teacher demonstrations to show how to formulate \colgrep queries, use retrieved evidence to guide subsequent actions, and produce correctly formatted submissions. We study two ways of using relevance labels during SFT: trajectory selection and turn-level weighting.

\paragraph{Teacher demonstration generation.}
We use DeepSeek-V4-Flash-0731 \citep{deepseekai2026deepseekv4} to generate the training instances described in Section~\ref{sec:training} through the localization harness. Trajectories are generated with temperature 0.3, at most 10 turns, 5 parallel tool calls per turn, and 2,048 completion tokens per turn. We slightly adapt the prompt at generation time to encourage the Teacher to use \colgrep rather than \grep. Furthermore, compatible \grep calls are replaced with regex-filtered \colgrep calls using the \emph{pattern} argument, while incompatible shell searches are resampled. Before training, we restore the prompt and tool schema used at evaluation.

\paragraph{Filtering and turn-level credit assignment.} We compare several trajectory-selection variants. \emph{Unfiltered} retains all valid teacher trajectories, while \emph{F1 filtering} requires a submitted-item F1 score of at least 0.5, at least one search action, and a \texttt{finish} submission within the interaction budget. We also revisit quality-weighted supervision \citep{qin2025supervised} through \emph{weighted SFT}, assigning credit to assistant turns based on their tool outcomes. We use the following weight assignment: discovering a new gold file receives weight 1.0, re-inspecting an already found gold file 0.6, retrieving only non-relevant files 0.3, and empty results, errors, or repeated commands 0.05. The final submission turn is weighted by its recall over the gold files. Prompt and tool-observation tokens are masked from the loss, and gold annotations are used only to construct training weights and are unavailable at evaluation. Let $\mathcal{T}_i$ be the assistant-token positions in turn $i$, $w_i$ its credit, and $\operatorname{CE}_t$ the token-level cross-entropy loss. We minimize,
\begin{equation}
\mathcal{L}_{\mathrm{SFT}}
=
\frac{
\sum_i w_i \sum_{t \in \mathcal{T}_i} \operatorname{CE}_t
}{
\sum_i w_i \lvert \mathcal{T}_i \rvert
}.
\label{eq:sft}
\end{equation}

\paragraph{Training configuration.}
We perform a separate full fine-tune for each student and supervision variant using TRL \citep{vonwerra2020trl}. We use AdamW \citep{loshchilov2017decoupled} with a learning rate of $2\times10^{-5}$, cosine decay, 3\% warmup, zero weight decay, gradient clipping at 1.0, bf16 precision, FlashAttention-3 \citep{shah2024flashattention}, sequence packing, and gradient checkpointing for one epoch.

\subsection{Reinforcement Learning for Efficient Localization}
\label{sec:rl}

Reinforcement learning (RL) allows the student to refine its use of \colgrep through direct interaction with the search environment. Starting from an SFT checkpoint, the student explores alternative queries and inspection strategies and is rewarded according to the quality of its final submission. This enables it to develop search behavior beyond the teacher demonstrations.

\paragraph{RL rewards and turn penalty.}
We compare multiple reward functions for code localization. \emph{File F1} rewards a balance between retrieving the correct files and avoiding incorrect ones, while \emph{File $F_2$} places more weight on recall, encouraging the agent to recover as many gold files as possible. Finally, \emph{File+Class+Function F1}, following CodeScout \citep{sutawika2026codescout}, sums localization scores at the file, class, and function levels to provide finer-grained supervision. We use this combined reward for Python instances, where class and function annotations are available, and File F1 alone for other languages. All rewards also include a format penalty of $-0.5$ when a rollout performs no search action or fails to submit through the \texttt{finish} tool, encouraging the agent to gather evidence and return a structured prediction. We also conduct an efficiency-focused file-F1 run, where we apply a bounded penalty based on the number of assistant turns $T$. The first two turns are unpenalized; each additional turn reduces the task reward by 1\%, up to a maximum reduction of 8\%. Here, \(R_{\mathrm{task}}\) denotes the task reward, \(R_{\mathrm{format}}\) the format penalty, \(T\) the number of assistant turns, \(T_0\) the unpenalized turn threshold, \(\lambda\) the per-turn penalty rate, and \(p_{\max}\) the maximum penalty.
\begin{equation}
R =
R_{\mathrm{task}}
\left[
1 -
\min\left(
p_{\max},
\lambda \max(0, T - T_0)
\right)
\right]
+
R_{\mathrm{format}}
\end{equation}

\paragraph{Training configuration.}
We use Prime-RL \citep{primeintellect2025prime-rl} to train the student with GRPO \citep{shao2024deepseekmath}, with several modifications: we remove the KL penalty and standard-deviation normalization of group-relative advantages, following Dr.~GRPO \citep{liu2025understanding}, and apply zero-advantage filtering with dynamic sampling, following DAPO \citep{yu2026dapo}. Each batch contains eight groups of 16 rollouts, totaling 128 rollouts, sampled at temperature 1.0 with at most 1,024 completion tokens per turn. Rollouts are generated asynchronously and discarded when they are more than four policy updates behind the current student. We update all model parameters for 500 steps with AdamW at a learning rate of $10^{-6}$, using two H100 GPUs for training and six for inference.

\begin{table}[!htbp]
\centering
\small
\setlength{\tabcolsep}{5pt}
\begin{tabular}{l rrr}
\toprule
& \multicolumn{3}{c}{\textbf{Code}}\\
\cmidrule(lr){2-4}
\textbf{Configuration}
& \textbf{SWE F1}
& \textbf{Multi-SWE F1}
& \textbf{Avg.\ retrieval calls}\\
\midrule
BM25 oracle, direct retrieval (top-10)
& 53.67 & 34.77 & 1.00\\
\colgrep oracle, direct retrieval (top-10)
& 59.67 & 46.79 & 1.00\\
Qwen3-1.7B RL, single-turn
& 61.86 & 51.25 & 2.00\\
Qwen3-1.7B RL, multi-turn
& \textbf{67.34} & \textbf{53.37} & 2.06\\
\bottomrule
\end{tabular}
\caption{File F1 (\%) and average retrieval calls for direct top-10 retrieval with oracle selection and single- and multi-turn Qwen3-1.7B RL agents on SWE-bench Lite and Multi-SWE-bench Flash.}
\label{tab:direct_single_multi}
\vspace{-0.5em}
\end{table}

\section{Results and Analysis}

\subsection{Training Effects on Search Interfaces}
\label{sec:interfaces}

We compare semantic search with \colgrep and lexical search by constructing a matched \grep variant of the agent harness. We remove the \colgrep tool, adjust the system prompt (Appendix \ref{sec:harness_prompts}) to describe the remaining lexical-search interface, and regenerate teacher trajectories with DeepSeek-V4-Flash-0731 under this modified harness. We then apply the same weighted-SFT and RL pipeline to these trajectories. This makes the comparison symmetric in the sense that each agent is trained on demonstrations generated with the tool available to it, while preserving the remaining training and evaluation protocol.

Table~\ref{tab:qwen_grep_vs_colgrep_training} shows that \colgrep outperforms \grep on file F1, our primary metric, on both benchmarks at every training stage. This advantage also holds across all three model families after weighted SFT (Table~\ref{tab:grep_vs_colgrep_full}). These results suggest that semantic retrieval reduces the search burden on small policies: the agent specifies the desired behavior, while the retriever matches and ranks candidate code. With \grep, by contrast, the policy must infer repository-specific terms and iteratively refine its queries based on returned matches. The training dynamics support this interpretation. Under the turn penalty, \colgrep-trained models converge to only a few search calls, suggesting that these are sufficient for file selection, whereas \grep-trained models continue to use more turns and tool calls (Figure~\ref{fig:rl-training-evolution}). At finer granularity, however, the \grep-based Qwen3-1.7B achieves higher class and function F1 after SFT and RL on SWE-bench Lite. This may reflect the value of exact lexical matching for distinguishing specific symbols within relevant files. Although \colgrep supports analogous filtering through its \texttt{pattern} argument, the student rarely uses it. Training also substantially narrows the interface gap: on Multi-SWE-bench Flash, \colgrep's file-F1 advantage decreases from 17.01 points for the base policy to 11.74 after weighted SFT and 3.84 after RL. Qualitative examples of these different search behaviors are provided in Appendix~\ref{sec:trajectory-examples}.

Figure~\ref{fig:rl-training-evolution} shows that, as RL progresses, the Qwen3-1.7B policy with \colgrep converges to much shorter search trajectories. After a few hundred updates, rollouts average roughly two \colgrep calls and three assistant turns, suggesting that only a few semantic searches are needed to identify relevant files before inspecting candidates and submitting the final answer.

\subsection{What Does The Agent Add Beyond Direct Retrieval?}

To isolate the contribution of agentic search beyond a single retrieval call, we compare our policy with \colgrep and BM25 direct-retrieval baselines. Each baseline uses the full original issue text as a single query and returns an oracle subset of the top-10 retrieved files. Because this subset is selected using the gold files, it represents an upper bound for that fixed retrieval call rather than a deployable system. We evaluate the Qwen3-1.7B policy trained with file-F1 reward and a turn penalty in two settings. In the single-turn setting, the model executes only its first parallel batch of tool calls, primarily \colgrep queries, and then submits immediately. In the multi-turn setting, it follows the evaluation protocol from Section~\ref{sec:eval}, with a maximum of 10 turns, though the model's average interaction length is only 3.7 turns.

As shown in Table~\ref{tab:direct_single_multi}, the single-turn agent outperforms both direct-retrieval baselines, exceeding BM25 by 8.19 file-F1 points on SWE-bench Lite and 16.48 points on Multi-SWE-bench Flash, and the stronger \colgrep baseline by 2.19 and 4.46 points, respectively. The agent formulates shorter, more targeted \colgrep queries and averages two retrieval calls in its first turn, allowing different queries to retrieve complementary evidence. Multi-turn interaction provides further gains, reaching 67.34 and 53.37 file F1, respectively. These gains require little additional retrieval: the multi-turn agent averages 2.06 \colgrep calls, compared with 2.00 for the single-turn variant. The additional interaction steps allow the model to verify and refine retrieved candidates, often using \texttt{sed} commands to inspect file contents before making its final prediction.

\begin{table*}[!tbp]
\centering\small
\setlength{\tabcolsep}{3pt}
\renewcommand{\arraystretch}{1.05}

\newcolumntype{M}{>{\centering\arraybackslash}p{0.75cm}}
\newcolumntype{B}{>{\columncolor{cyan!10}\centering\arraybackslash}p{0.75cm}}

\resizebox{\textwidth}{!}{%
\begin{tabular}{ll B M M M M B M M}
\toprule
& &
\multicolumn{5}{c}{\textbf{SWE-bench Lite}} &
\multicolumn{3}{c}{\textbf{Multi-SWE-bench Flash}} \\
\cmidrule(lr){3-7}
\cmidrule(lr){8-10}

& &
\multicolumn{3}{c}{\textbf{File}} &
\multicolumn{1}{c}{\textbf{Class}} &
\multicolumn{1}{c}{\textbf{Function}} &
\multicolumn{3}{c}{\textbf{File}} \\
\cmidrule(lr){3-5}
\cmidrule(lr){6-6}
\cmidrule(lr){7-7}
\cmidrule(lr){8-10}

\multicolumn{1}{c}{\textbf{Student}} &
\multicolumn{1}{c}{\textbf{Method/Filtering}} &
\cellcolor{cyan!10}\textbf{F1} & R & P & F1 & F1 &
\cellcolor{cyan!10}\textbf{F1} & R & P \\
\midrule

Qwen3-1.7B & Base
& 34.51 & 56.53 & 29.30 & 13.64 & 2.81
& 29.39 & 43.18 & 31.45 \\
& Unfiltered
& 54.87 & 71.33 & 48.08 & 27.69 & 7.58
& 43.85 & 53.27 & 46.82 \\
& F1
& 53.50 & 68.53 & 47.31 & 25.14 & 6.70
& 42.15 & 49.40 & 46.13 \\
& Weighted
& \textbf{55.09} & \textbf{71.67} & \textbf{48.46}
& \textbf{28.62} & \textbf{8.00}
& \textbf{44.58} & \textbf{54.26} & \textbf{47.31} \\

\midrule

MiniCPM5-1B & Base
& 7.72 & 20.93 & 5.67 & 2.69 & 0.47
& 7.50 & 13.93 & 7.27 \\
& Unfiltered
& 45.23 & 59.47 & 39.62 & 20.18 & \textbf{3.26}
& 38.06 & 45.48 & 43.00 \\
& F1
& 43.85 & 55.80 & 39.00 & 18.92 & 2.79
& 36.80 & 41.90 & 42.74 \\
& Weighted
& \textbf{47.10} & \textbf{60.53} & \textbf{41.77}
& \textbf{20.93} & 2.52
& \textbf{40.77} & \textbf{47.84} & \textbf{45.84} \\

\midrule

LFM2.5-1.2B-Instruct & Base
& 11.29 & 18.13 & 9.04 & 0.85 & 0.32
& 6.75 & 8.00 & 7.80 \\
& Unfiltered
& 45.71 & 59.07 & 40.36 & \textbf{21.24} & 4.09
& 37.79 & 45.21 & 41.89 \\
& F1
& 43.98 & 55.47 & 39.20 & 18.07 & 3.02
& 35.50 & 41.87 & 40.16 \\
& Weighted
& \textbf{46.87} & \textbf{60.73} & \textbf{41.34}
& 20.97 & \textbf{4.20}
& \textbf{39.06} & \textbf{47.08} & \textbf{42.04} \\

\midrule
DeepSeek-V4-Flash-0731 & -
  & 63.96 & 90.73 & 53.26 & 50.30 & 30.51
  & 59.12 & 74.85 & 57.34 \\

\bottomrule
\end{tabular}
}

\caption{\textbf{Comparison of supervision filtering methods} (unfiltered, F1, and turn-weighted) across three model families on code localization. Bold denotes the best result for each student and metric.}
\label{tab:sft}
\end{table*}

\subsection{How Do Supervision and Reward Choices Affect the Policy?}

Table~\ref{tab:sft} shows that turn-weighted SFT achieves the highest file F1 for all three students on both benchmarks. This supports the value of turn-level supervision for multi-turn search: unsuccessful trajectories can still contain useful retrieval steps, while successful ones may include unproductive turns. Weighting each turn by its observed retrieval outcome preserves informative behavior without assigning equal importance to every action. We therefore use the weighted SFT checkpoint to initialize all subsequent RL experiments.

RL reward design reveals distinct trade-offs (Table~\ref{tab:rewards}). Relative to weighted SFT, optimizing file F1 improves file F1 by 12.20 points on SWE-bench Lite and 9.42 points on Multi-SWE-bench Flash. Optimizing $F_2$ achieves the highest recall on both benchmarks but reduces file F1, while the combined file, class, and function reward yields the highest class and function F1 on SWE-bench Lite. Adding the turn penalty keeps file F1 within 0.63 points of the unpenalized file-F1 reward while producing much shorter trajectories.

\subsection{Search Behavior and Measured Efficiency}

\begin{figure}[!tbp]
    \centering
    \includegraphics[width=0.9\textwidth]{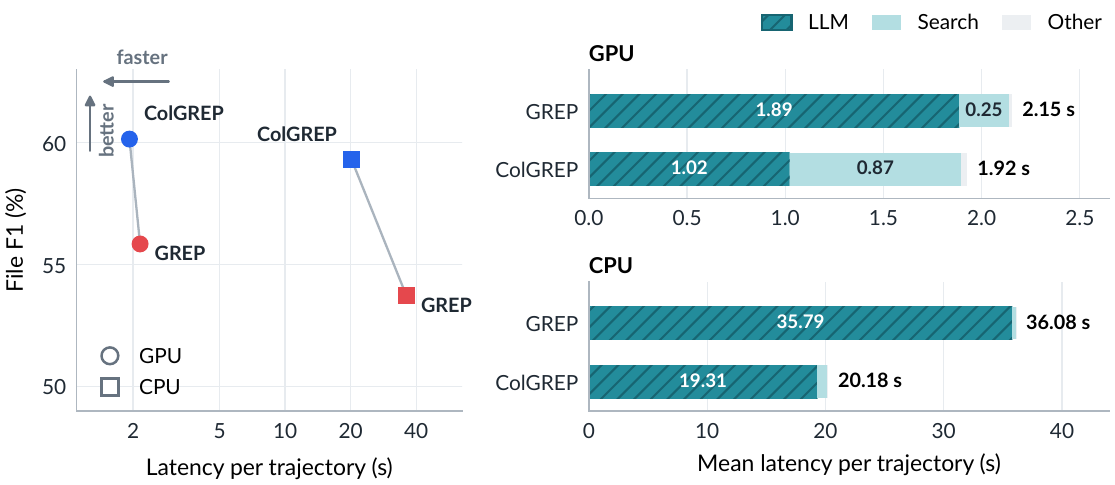}
    \caption{\textbf{File-localization performance and latency for Qwen3-1.7B after RL} with the turn penalty. The latency breakdown separates LLM inference, search (\grep /\colgrep calls), and other runtime overhead.}
    \label{fig:rl-search-behavior}

\end{figure}

We study the search behavior and efficiency of the \grep- and \colgrep-based policies after RL training with file-F1 reward and our turn penalty. We compare trajectory length, end-to-end latency, and latency composition to understand how the search interface affects both localization accuracy and inference cost. For latency measurements, trajectories are executed sequentially in both GPU and CPU settings, with the student model served through vLLM on GPU or llama.cpp with BF16 GGUF weights on CPU. In both settings, \colgrep runs entirely on CPU over precomputed indexes; see Appendix~\ref{app:latency-protocol} for the full measurement protocol.

\paragraph{End-to-end efficiency.} Averaged equally over the two benchmarks, \colgrep is substantially more efficient than \grep, with the largest gains appearing in the CPU-only setting. On CPU, mean end-to-end latency decreases from 36.08 to 20.18 seconds per trajectory, a reduction of 44.1\%. With GPU inference, latency also decreases, from 2.15 to 1.92 seconds, corresponding to a 10.6\% reduction. Trajectories are also considerably shorter, with mean length decreasing from 7,857 to 5,574 tokens, a reduction of 29.1\%. Together with the higher file F1 shown in Figure~\ref{fig:rl-search-behavior}, these results indicate that \colgrep reaches better predictions with both fewer tokens and lower end-to-end latency.

\paragraph{Latency composition.} We further examine how end-to-end latency is distributed between LLM inference and search calls, i.e., time spent executing \grep or \colgrep. With GPU inference, LLM inference accounts for 87.6\% of \grep latency, compared with 53.1\% for \colgrep. \colgrep cuts LLM inference time nearly in half, which more than compensates for the higher cost of \colgrep calls, reducing total latency from 2.15 to 1.92 seconds. This difference becomes much more pronounced on CPU, where LLM inference accounts for 99.2\% of \grep latency, compared with 95.7\% for \colgrep. Since \colgrep requires substantially less LLM inference, the end-to-end latency reduction grows from 10.6\% with GPU inference to 44.1\% on CPU. The reported \colgrep latency also includes avoidable retrieval overhead, since each call currently launches a new process and reloads the retrieval model; keeping the model resident across calls could further reduce latency.

\subsection{Generalization to Unseen Programming Languages}
\label{sec:language}

\begin{figure}[!tbp]
    \centering
    \includegraphics[width=0.9\textwidth]{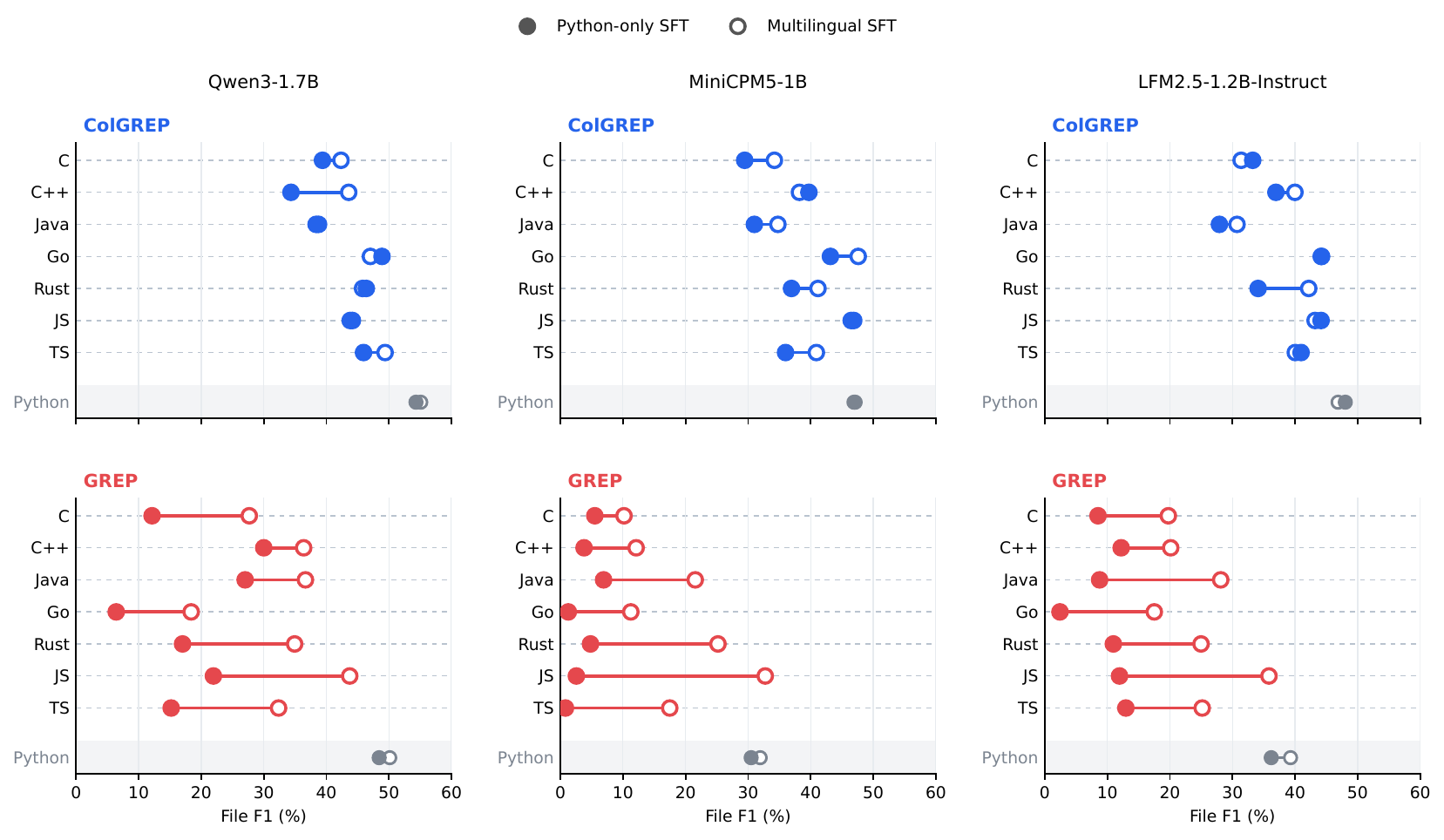}
    \caption{Performance across seven programming languages after Python-only versus multilingual SFT, using the \colgrep and \grep harnesses.}
    \label{fig:language generalization}
    \vspace{-1em}
\end{figure}

While our training recipe uses multilingual data for both SFT and RL, we also study SFT restricted to a single programming language. Specifically, we fine-tune each student with weighted SFT on Python-only trajectories generated by DeepSeek-V4-Flash-0731. For each setup reported in Table~\ref{tab:sft_languages}, we collect additional decontaminated Python-only teacher trajectories to match the sample count used for multilingual weighted SFT. We then evaluate on the seven non-Python languages in Multi-SWE-bench Flash and compare these policies with their multilingual weighted-SFT counterparts, isolating the effect of excluding evaluation languages from SFT while holding the training sample count fixed.

Figure~\ref{fig:language generalization} shows that restricting SFT to Python has a substantially smaller effect on \colgrep-based policies. Averaged across the three students, benchmark file F1 drops by only 2.13 points with \colgrep, compared with 14.90 points with \grep. This pattern holds for all three students: the benchmark-level drop ranges from 1.47 to 3.04 points with \colgrep, versus 14.53 to 15.27 points with \grep. Under Python-only training, \colgrep also outperforms \grep across all seven languages and all three students, with an average benchmark advantage of 28.82 points. These results suggest that, in our setting, a multilingual code retriever substantially reduces the dependence on multilingual policy fine-tuning. The agent formulates natural-language queries, while the retriever matches them to code across languages. With \grep, the agent must instead generate search patterns that account for language-specific syntax and identifiers.

\section{Discussion and Limitations}
\label{sec:limitations}

While this work shows that compact agents benefit from semantic repository search, several limitations remain. First, we restrict our study to models below two billion parameters, reflecting our focus on fast, inexpensive agents that can run locally on consumer hardware. Evaluating larger models would help determine whether these benefits persist as the underlying policy becomes more capable, but is beyond the scope of this work. We also conduct all experiments using a single agent harness and therefore do not assess the sensitivity of our findings to different agent implementations, prompting strategies, or tool interfaces. Finally, our evaluation measures localization against files modified by a reference patch, which captures only one possible solution and may omit useful contextual files. Our experiments are also limited to code repositories. Extending the framework to other agent harnesses and retrieval-intensive domains, such as general document search, remains an important direction for future work.

\section{Conclusion}

We presented a training framework for compact code-localization agents combining weighted supervised fine-tuning with reinforcement learning around \colgrep. Across SWE-bench Lite and Multi-SWE-bench Flash, semantic search improves localization over lexical search, while weighted supervision improves file F1 across three model families below two billion parameters. Semantic search also reduces trajectory length by 29.1\% and mean latency by 10.6\% on GPU and 44.1\% on CPU, supporting resource-constrained deployment. Our analyses further show stronger generalization to unseen programming languages and that multi-turn interaction outperforms both single-turn search and direct retrieval. Overall, these results support compact semantic-search agents as an efficient approach to multilingual file localization.

\subsection*{AI use statement}

In this work, we used generative AI tools as part of our methodology, in particular to generate synthetic training data with DeepSeek, as described in Section \ref{sec:training-sft-rl}. We also used AI assistants to support the preparation of the manuscript, including proofreading and polishing the writing, improving clarity and phrasing, and suggesting possible section titles; all final titles and formulations were selected and reviewed by the authors. Where AI-assisted outputs were used in the research workflow or manuscript preparation, they were reviewed by the authors before inclusion. We did not rely on generative AI to determine the scientific conclusions of the work. We take full responsibility for the final content of this work, including all text, claims, analyses, and artifacts produced with the assistance of generative AI.

\subsection*{Ethics statement}

This work studies code localization using public software repositories and benchmarks. It does not involve human participants, personal information, or sensitive data. We do not identify any specific ethical concerns associated with the methods or experiments presented in this paper.

\subsection*{Reproducibility statement}

We describe the evaluation benchmarks and metrics in Section~\ref{sec:eval}, the construction and decontamination of the training data in Section~\ref{sec:training}, and the agent harness, retrieval configuration, and training procedure in Sections~\ref{sec:harness} and ~\ref{sec:training-sft-rl}. 

% To support reproducibility, we will release model checkpoints, software, data, and training pipelines upon acceptance.

\bibliographystyle{plainnat}
\bibliography{iclr2027_conference}

\clearpage

\appendix

\input{iclr2027_appendix}

\end{document}

%% file: math_commands.tex
\usepackage{amsmath,amsfonts,bm}

\def\eqref#1{equation~\ref{#1}}
\def\1{\bm{1}}

\DeclareMathAlphabet{\mathsfit}{\encodingdefault}{\sfdefault}{m}{sl}
\SetMathAlphabet{\mathsfit}{bold}{\encodingdefault}{\sfdefault}{bx}{n}

%% file: figures/overview_codeonly.tex
% Duplicate of secondary path: explicit repository and pre-built index. Drawn at 13.6 cm publication width.
% Turn weights are schematic; language icons show the union across training stages.
\begingroup%
\definecolor{dzred}{HTML}{E5484D}%
\definecolor{dzblue}{HTML}{2563EB}%
\definecolor{dzpurple}{HTML}{7954A0}%
\definecolor{dzpale}{HTML}{EEE5F7}%
\definecolor{dzsage}{HTML}{47776D}%
\definecolor{dzsagepale}{HTML}{EDF4F1}%
\definecolor{dzamber}{HTML}{9A6B2E}%
\definecolor{dzamberpale}{HTML}{FBF0DC}%
\input{figures/assets/overview-icons}%
\begin{tikzpicture}[paper figure,
  % Three text levels at publication size: 9 pt headings, 8 pt titles,
  % 7 pt annotations; literal paths and commands use 6.5 pt monospace.
  dz heading/.style={font=\fontsize{9}{10.5}\selectfont\bfseries,
    text=dzpurple,anchor=west,inner sep=0pt},
  dz title/.style={font=\fontsize{8}{9.5}\selectfont\bfseries,
    text height=1.5ex,text depth=.25ex,inner sep=0pt,align=center},
  dz note/.style={font=\fontsize{7}{8.5}\selectfont,
    text=figmuted,align=center,inner sep=0pt},
  dz small/.style={dz note,text height=1.5ex,text depth=.25ex},
  dz code/.style={dz small,font=\ttfamily\fontsize{6.5}{8}\selectfont},
  dz flow/.style={fig arrow,line width=1pt},
  dz box/.style={fig box,inner sep=0pt}]
\path[use as bounding box] (0,-.44) rectangle (13.6,8.65);

\begin{scope}[yshift=1.00cm]
\begin{scope}[yshift=-.77cm]
% A single multilingual source band, without claiming identical SFT/RL sets.
\filldraw[draw=dzsage!55,fill=dzsagepale,rounded corners=3pt]
  (.04,6.85) rectangle (13.56,8.38);
\node[dz small,text=dzsage] at (6.8,8.03)
  {\includegraphics[height=.17cm]{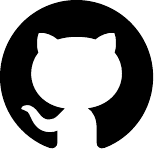}\hspace{.08cm}%
   \textbf{Training data:} issue--patch pairs};
\node[inner sep=0pt] at (.84,7.53)
  {\includegraphics[width=.35cm,height=.35cm,keepaspectratio]{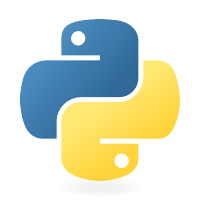}};
\node[dz small,text=figink] at (.84,7.19) {Python};

\node[inner sep=0pt] at (2.33,7.53)
  {\includegraphics[width=.35cm,height=.35cm,keepaspectratio]{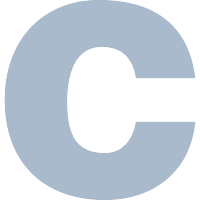}};
\node[dz small,text=figink] at (2.33,7.19) {C};

\node[inner sep=0pt] at (3.82,7.53)
  {\includegraphics[width=.35cm,height=.35cm,keepaspectratio]{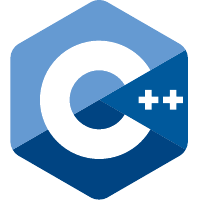}};
\node[dz small,text=figink] at (3.82,7.19) {C++};

\node[inner sep=0pt] at (5.31,7.53)
  {\includegraphics[width=.35cm,height=.35cm,keepaspectratio]{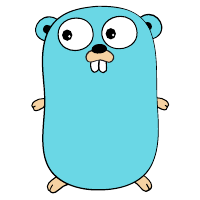}};
\node[dz small,text=figink] at (5.31,7.19) {Go};

\node[inner sep=0pt] at (6.80,7.53)
  {\includegraphics[width=.35cm,height=.35cm,keepaspectratio]{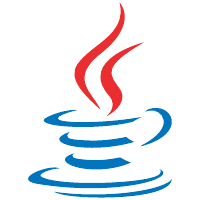}};
\node[dz small,text=figink] at (6.80,7.19) {Java};

\node[inner sep=0pt] at (8.29,7.53)
  {\includegraphics[width=.35cm,height=.35cm,keepaspectratio]{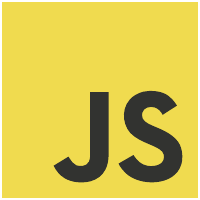}};
\node[dz small,text=figink] at (8.29,7.19) {JavaScript};

\node[inner sep=0pt] at (9.78,7.53)
  {\includegraphics[width=.35cm,height=.35cm,keepaspectratio]{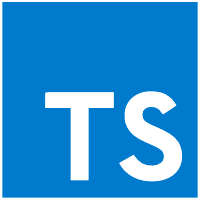}};
\node[dz small,text=figink] at (9.78,7.19) {TypeScript};

\node[inner sep=0pt] at (11.27,7.53)
  {\includegraphics[width=.35cm,height=.35cm,keepaspectratio]{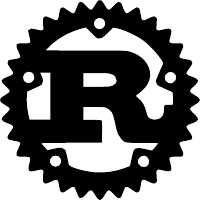}};
\node[dz small,text=figink] at (11.27,7.19) {Rust};

\node[inner sep=0pt] at (12.76,7.53)
  {\includegraphics[width=.35cm,height=.35cm,keepaspectratio]{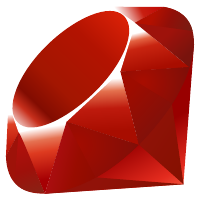}};
\node[dz small,text=figink] at (12.76,7.19) {Ruby};

% Stage labels describe the colgrep training recipe.
\node[dz title] at (1.98,6.47) {Generate};
\node[dz title] at (1.98,6.18) {Teacher traces};
\node[dz title] at (5.90,6.47) {Weighted SFT};
\node[dz title] at (9.00,6.47) {RL};
\node[dz title] at (12.10,6.47) {Trained agent};
\node[dz small] at (5.90,6.18) {Select traces; weight turns};
\node[dz small] at (9.00,6.18) {Search $\to$ submit $\to$ reward};
\node[dz small] at (12.10,6.18) {1B--1.7B parameters};

% Teacher traces train a compact policy specialized for colgrep.
\node[inner sep=0pt] at (.96,5.52)
  {\includegraphics[width=1.65cm]{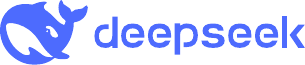}};
\foreach \variant/\tone/\y/\search/\credits in {
  colgrep/dzblue/5.52/{Semantic search}/{-.45/.096/35,-.15/.32/90,.15/.016/15,.45/.192/60}}{
  % A pale ribbon and colored arrows carry each variant through all stages.
  \fill[\tone!4,rounded corners=5pt] (2.13,\y-.46) rectangle (13.56,\y+.46);
  \draw[dz flow,draw=\tone] (1.82,\y)-- (2.24,\y);
  % Offset sheets identify a dataset of trajectories rather than a search tool.
  \node[dz box,draw=\tone!25,fill=white,minimum width=1.67cm,
    minimum height=.65cm] at (3.09,\y+.075) {};
  \node[dz box,draw=\tone!65,fill=white,minimum width=1.67cm,
    minimum height=.65cm] (traces-\variant) at (3.01,\y) {};
  \node[dz title,text=\tone] at (3.01,\y+.11) {\csname\variant\endcsname};
  \node[dz small,text=\tone] at (3.01,\y-.15) {traces};
  % Outcome-weighted assistant turns teach the compact student.
  \node[dz box,draw=\tone!40,fill=white,minimum width=2.70cm,
    minimum height=.70cm] (sft-\variant) at (5.90,\y) {};
  \begin{scope}[shift={(5.25,\y)},draw=\tone]
    \draw[draw=\tone!30] (-.61,-.16)--(.61,-.16);
    \foreach \x/\height/\shade in \credits {
      \filldraw[draw=\tone!65,fill=\tone!\shade!white,rounded corners=.25pt]
        (\x-.10,-.16) rectangle (\x+.10,-.16+\height);
    }
  \end{scope}
  \draw[dz flow,draw=\tone,line width=.65pt] (5.96,\y)--(6.25,\y);
  \begin{scope}[shift={(6.69,\y)},x=.39cm,y=.39cm,draw=\tone,
    fill=\tone,line width=.8pt,line cap=round,line join=round]
    \pic {overview-student};
  \end{scope}
  % RL refines the checkpoint with environment feedback.
  \filldraw[fill=white,draw=\tone!30] (9.00,\y) circle (.43);
  \begin{scope}[shift={(9.00,\y)},x=.87cm,y=.87cm,draw=\tone,
    line width=.85pt,line cap=round]
    \pic {overview-learn};
  \end{scope}
  \begin{scope}[shift={(9.00,\y)},x=.27cm,y=.27cm,draw=\tone,
    fill=\tone,line width=.6pt,line cap=round,line join=round]
    \pic {overview-student};
  \end{scope}
  % A saturated policy badge marks the trained agent.
  \node[dz box,draw=\tone,fill=\tone,rounded corners=4pt,
    minimum width=2.70cm,minimum height=.79cm] (policy-\variant) at (12.10,\y) {};
  \begin{scope}[shift={(11.24,\y)},x=.43cm,y=.43cm,draw=white,
    fill=white,line width=.8pt,line cap=round,line join=round]
    \pic {overview-student};
  \end{scope}
  \node[dz title,text=white]
    at (12.46,\y+.09) {\csname\variant\endcsname};
  \node[dz small,text=white] at (12.46,\y-.19) {trained agent};
  \draw[dz flow,draw=\tone] (traces-\variant.east)--(sft-\variant.west);
  \draw[dz flow,draw=\tone] (sft-\variant.east)--(8.57,\y);
  \draw[dz flow,draw=\tone] (9.43,\y)--(policy-\variant.west);
}

\end{scope}

% A thin secondary path exposes the comparison without a second large ribbon.
\node[dz note,text=dzred,align=left,anchor=west] at (.10,3.62)
  {\textbf{Lexical}\\\textbf{comparison}};
\draw[dz flow,draw=dzred!65,line width=.6pt]
  (1.82,4.75) to[out=-70,in=180] (2.40,3.66);
\node[dz title,text=dzred] (baseline-traces) at (3.01,3.76) {\grep};
\node[dz small] at (3.01,3.48) {separate traces};
\foreach \x/\label in {5.90/Weighted SFT,9.00/RL}{
  \fill[dzred!75] (\x,3.66) circle (.045);
  \node[dz small,above=4pt] at (\x,3.66) {\label};
}
\draw[dz flow,draw=dzred!65,line width=.6pt] (3.77,3.66)--(5.70,3.66);
\draw[dz flow,draw=dzred!65,line width=.6pt] (6.10,3.66)--(8.80,3.66);
\draw[dz flow,draw=dzred!65,line width=.6pt] (9.20,3.66)--(10.92,3.66);
\begin{scope}[shift={(11.24,3.66)},x=.31cm,y=.31cm,draw=dzred,
 fill=dzred,line width=.65pt,line cap=round,line join=round]
 \pic {overview-student};
\end{scope}
\node[dz title,text=dzred] at (12.30,3.77) {\grep};
\node[dz small] at (12.30,3.48) {trained agent};

\end{scope}

% The task example is illustrative; the index is prepared before agent search.
\draw[draw=figline] (.04,4.18)--(13.56,4.18);
\node[dz heading] at (.04,3.90) {Localize files for an issue};

% A repository tree makes the search space visible alongside the issue description.
\node[dz box,draw=dzsage!65,fill=dzsagepale,minimum width=3.65cm,
  minimum height=2.28cm] (repo) at (1.89,2.51) {};
\node[inner sep=0pt] at (.43,3.40)
  {\includegraphics[height=.27cm]{figures/assets/github-mark.pdf}};
\node[dz small,text=dzsage,anchor=west] at (.71,3.40) {Issue};
\node[dz note,text=figink,align=left,anchor=west] at (.28,2.96)
  {``Support custom delimiters\\in CSV imports.''};
\draw[draw=dzsage!25] (.28,2.62)--(3.50,2.62);
\node[inner sep=0pt] at (.43,2.37)
  {\includegraphics[height=.27cm]{figures/assets/github-mark.pdf}};
\node[dz small,anchor=west,text=dzsage] at (.71,2.37) {Repository};
\draw[draw=dzsage!55,line width=.5pt] (.43,2.16)--(.43,1.58);
\foreach \y/\folder in {2.08/{src/csv/},1.83/{src/utils/},1.58/{tests/}}{
  \draw[draw=dzsage!55,line width=.5pt] (.43,\y)--(.63,\y);
  \node[dz code,anchor=west,text=figink]
    at (.72,\y) {\folder};
}
\foreach \x/\y in {2.50/2.08,2.83/1.93,3.16/1.78}{
  \begin{scope}[shift={(\x,\y)},x=.24cm,y=.24cm,draw=dzsage!70,
    fill=white,line width=.55pt,line cap=round,line join=round]
    \pic {overview-document};
  \end{scope}
}

% Candidate files are returned; neither a patch nor a verified error is implied.
\node[dz box,draw=dzamber!65,fill=dzamberpale,minimum width=3.65cm,
  minimum height=1.64cm] (files) at (11.71,2.65) {};
\node[dz title,anchor=west,text=dzamber] at (10.12,3.10) {Candidate files};
\foreach \y/\file in {2.65/{src/csv/parser.py},2.22/{src/csv/reader.py}}{
  \begin{scope}[shift={(10.29,\y)},x=.25cm,y=.25cm,draw=dzamber,
    line width=.65pt,line cap=round,line join=round]
    \pic {overview-files};
  \end{scope}
  \node[dz code,anchor=west,text=figink]
    at (10.59,\y) {\file};
}
\node[dz note,anchor=west,align=left] at (files.west |- 0,1.40)
  {Evaluation\\metric};
\node[dz box,draw=dzamber!65,fill=dzamberpale,minimum width=1.70cm,
  minimum height=.58cm,anchor=east] (metric) at (files.east |- 0,1.40) {};
\node[dz title,text=dzamber] at (metric.center) {File F1};

\node[dz box,draw=dzpurple,fill=dzpale,line width=1pt,
  minimum width=3.85cm,minimum height=1.06cm] (agent) at (6.8,2.65) {};
\fill[dzpurple] (5.57,2.65) circle (.34);
\begin{scope}[shift={(5.57,2.65)},x=.37cm,y=.37cm,draw=white,
  fill=white,line width=.7pt,line cap=round,line join=round]
  \pic {overview-student};
\end{scope}
\node[dz title,text=dzpurple] at (7.20,2.65) {Trained student};
\draw[dz flow,draw=dzpurple] (repo.east |- agent.west)--(agent.west);
\draw[dz flow,draw=dzpurple] (agent.east)--
  node[dz small,above=3pt]{\texttt{finish}} (files.west);

% Each code unit has multiple token vectors, not a single pooled embedding.
% Matching belongs to colgrep; only code representations are stored in the index.
\node[dz box,draw=dzsage!65,fill=dzsagepale,minimum width=3.65cm,
  minimum height=.96cm] (index) at (1.89,.42) {};
% Two offset code-unit sheets, each containing a row of cells per token vector.
\foreach \dx/\dy in {.10/.10,0/0}{
  \begin{scope}[shift={(.32+\dx,.13+\dy)}]
    \filldraw[draw=dzsage!70,fill=white,rounded corners=.5pt]
      (0,0) rectangle (.65,.52);
    \foreach \y in {.08,.23,.38}{
      \foreach \x/\shade in {.08/75,.24/40,.40/60}{
        \fill[dzsage!\shade,rounded corners=.2pt]
          (\x,\y) rectangle +(.12,.08);
      }
    }
  \end{scope}
}
\node[dz title,text=dzsage,anchor=west] at (1.19,.60) {Pre-built index};
\node[dz small,text=dzsage,anchor=west] at (1.19,.25) {Code token vectors};
\draw[fig arrow,draw=dzsage!75,line width=.65pt]
  (repo.south)--node[dz small,right=4pt]{Before search}(index.north);

% All three tools are explicit; finish also labels the final submission arrow.
\node[dz box,draw=figline,fill=figgray,minimum width=5.30cm,
  minimum height=1.50cm] (tools) at (6.80,.335) {};
\node[dz box,draw=dzblue!60,fill=white,minimum width=4.80cm,
  minimum height=.46cm] (search) at (6.80,.72) {};
\node[dz small] at (6.80,.72)
  {\textcolor{dzblue}{\colgrep}\quad Late-interaction, semantic search};
\foreach \x/\tool/\description in {5.55/terminal/{Navigate / inspect},8.05/finish/{Submit files}}{
  \node[dz box,draw=figline,fill=white,minimum width=2.30cm,
    minimum height=.64cm] at (\x,.04) {};
  \node[dz code,text=figink]
    at (\x,.18) {\tool};
  \node[dz small]
    at (\x,-.10) {\description};
}
\draw[fig arrow,draw=dzblue,line width=.65pt]
  (search.west)--(index.east |- search.west);
\draw[dz flow,draw=dzpurple] ([xshift=-.82cm]agent.south)
  to[out=230,in=130] node[dz small,left=3pt]{calls}
  ([xshift=-.82cm]tools.north);
\draw[dz flow,draw=dzpurple] ([xshift=.82cm]tools.north)
  to[out=50,in=310] node[dz small,right=3pt]{results}
  ([xshift=.82cm]agent.south);
\end{tikzpicture}%
\endgroup%

%% file: iclr2027_appendix.tex
\section{Appendix}

\subsection{Supervised Fine-Tuning Data}
\label{app:sft-data}

Table~\ref{tab:sft_languages} provides additional details on the supervised fine-tuning data used in our experiments. In particular, we report the language composition of the teacher trajectories retained under the different filtering setups

\begin{table}[!htbp]
    \centering
    \small
    \begin{tabular}{lrrrr}
        \toprule
        & \grep & \multicolumn{3}{c}{\colgrep} \\
        \cmidrule(lr){2-2}
        \cmidrule(lr){3-5}
        Language & Weighted & Unfiltered & Weighted & File-F1 $\geq 0.5$ \\
        \midrule
        Python     & 4,374 & 4,288 & 4,288 & 3,157 \\
        Go         & 2,056 & 2,083 & 2,083 & 1,402 \\
        Rust       & 1,840 & 1,841 & 1,841 & 1,291 \\
        JavaScript & 1,824 & 1,822 & 1,822 & 1,224 \\
        TypeScript & 1,697 & 1,684 & 1,684 & 1,056 \\
        Java       & 1,453 & 1,433 & 1,433 &   811 \\
        C          &   192 &   186 &   186 &   113 \\
        C++        &   114 &   115 &   115 &    70 \\
        \midrule
        Total      & 13,550 & 13,452 & 13,452 & 9,124 \\
        \bottomrule
    \end{tabular}
    \caption{
        Language distribution of teacher trajectories generated with
        DeepSeek-V4-Flash-0731 after cleaning and selection for each SFT mode.
    }
    \label{tab:sft_languages}
\end{table}

\subsection{Search Tool Comparison}
\label{app:search-tool-comparison}

Table~\ref{tab:grep_vs_colgrep_full} compares \grep and \colgrep after weighted supervised fine-tuning across three small language models. Across both SWE-bench Lite and Multi-SWE-bench Flash, \colgrep consistently improves file-level localization performance. The gains are particularly pronounced for the smaller LFM2.5-1.2B-Instruct and MiniCPM5-1B models.

\begin{table*}[!htbp]
    \centering
    \small
    \setlength{\tabcolsep}{3pt}
    \renewcommand{\arraystretch}{1.05}

    % Highlighted metric column
    \newcolumntype{B}{>{\columncolor{cyan!10}\centering\arraybackslash}r}

    \resizebox{\textwidth}{!}{%
        \begin{tabular}{lll B r r r r r}
            \toprule
            \textbf{Benchmark} & \textbf{Model} & \textbf{Search tool}
            & \multicolumn{3}{c}{\textbf{File}}
            & \multicolumn{1}{c}{\textbf{Class}}
            & \multicolumn{1}{c}{\textbf{Function}}
            & \textbf{Turns} \\
            \cmidrule(lr){4-6}
            \cmidrule(lr){7-7}
            \cmidrule(lr){8-8}

            & & &
            \cellcolor{cyan!10}\textbf{F1} & R & P &
            F1 & F1 & \\
            \midrule

            \multicolumn{9}{l}{\textbf{SWE-bench Lite}} \\

            & \multirow{2}{*}{Qwen3-1.7B + SFT}
            & \grep
            & 50.13 & 60.93 & 45.60
            & \textbf{33.58}
            & \textbf{18.00}
            & \textbf{7.89} \\
            & & \colgrep
            & \textbf{55.09} & \textbf{71.67} & \textbf{48.46}
            & 28.62
            & 8.00
            & 8.53 \\
            \cmidrule(lr){2-9}

            & \multirow{2}{*}{LFM2.5-1.2B-Instruct + SFT}
            & \grep
            & 39.28 & 47.33 & 36.16
            & \textbf{25.14}
            & \textbf{10.75}
            & \textbf{7.79} \\
            & & \colgrep
            & \textbf{46.87} & \textbf{60.73} & \textbf{41.34}
            & 20.97
            & 4.20
            & 7.83 \\
            \cmidrule(lr){2-9}

            & \multirow{2}{*}{MiniCPM5-1B + SFT}
            & \grep
            & 31.93 & 41.60 & 28.18
            & 20.20
            & \textbf{9.25}
            & 7.78 \\
            & & \colgrep
            & \textbf{47.10} & \textbf{60.53} & \textbf{41.77}
            & \textbf{20.93}
            & 2.52
            & \textbf{7.30} \\

            \midrule
            \multicolumn{9}{l}{\textbf{Multi-SWE-bench Flash}} \\

            & \multirow{2}{*}{Qwen3-1.7B + SFT}
            & \grep
            & 32.84 & 36.47 & 38.53
            & -- & --
            & \textbf{7.94} \\
            & & \colgrep
            & \textbf{44.58} & \textbf{54.26} & \textbf{47.31}
            & -- & --
            & 8.56 \\
            \cmidrule(lr){2-9}

            & \multirow{2}{*}{LFM2.5-1.2B-Instruct + SFT}
            & \grep
            & 24.59 & 26.87 & 30.11
            & -- & --
            & \textbf{7.88} \\
            & & \colgrep
            & \textbf{39.06} & \textbf{47.08} & \textbf{42.04}
            & -- & --
            & 7.91 \\
            \cmidrule(lr){2-9}

            & \multirow{2}{*}{MiniCPM5-1B + SFT}
            & \grep
            & 18.84 & 20.40 & 23.44
            & -- & --
            & 8.12 \\
            & & \colgrep
            & \textbf{40.77} & \textbf{47.84} & \textbf{45.84}
            & -- & --
            & \textbf{7.12} \\

            \bottomrule
        \end{tabular}%
    }
    \caption{
        Comparison of \grep and \colgrep after weighted SFT across three small
        language models. Bold indicates the better result for each model and
        metric.
    }
    \label{tab:grep_vs_colgrep_full}
\end{table*}

\subsection{Reinforcement Learning Reward Ablations}
\label{app:reward-ablations}

We next study how the choice of reward affects localization performance. Table~\ref{tab:rewards} compares file-level F1, an $F_2$ reward that places greater weight on recall, a hierarchical reward combining file-, class-, and function-level F1, and a file-F1 reward augmented with a turn penalty.

\begin{table*}[!htbp]
\centering
\small
\setlength{\tabcolsep}{3pt}
\renewcommand{\arraystretch}{1.05}

% If M and B are already defined earlier in the paper, omit these two lines.
\newcolumntype{M}{>{\centering\arraybackslash}p{0.75cm}}
\newcolumntype{B}{>{\columncolor{cyan!10}\centering\arraybackslash}p{0.75cm}}

\resizebox{\textwidth}{!}{%
\begin{tabular}{l B M M M M B M M}
\toprule
&
\multicolumn{5}{c}{\textbf{SWE-bench Lite}} &
\multicolumn{3}{c}{\textbf{Multi-SWE-bench Flash}} \\
\cmidrule(lr){2-6}
\cmidrule(lr){7-9}

&
\multicolumn{3}{c}{\textbf{File}} &
\multicolumn{1}{c}{\textbf{Class}} &
\multicolumn{1}{c}{\textbf{Function}} &
\multicolumn{3}{c}{\textbf{File}} \\
\cmidrule(lr){2-4}
\cmidrule(lr){5-5}
\cmidrule(lr){6-6}
\cmidrule(lr){7-9}

\multicolumn{1}{c}{\textbf{Training / reward}} &
\cellcolor{cyan!10}\textbf{F1} & R & P &
F1 & F1 &
\cellcolor{cyan!10}\textbf{F1} & R & P \\
\midrule

Weighted SFT
& 55.09 & 71.67 & 48.46
& 28.62 & 8.00
& 44.58 & 54.26 & 47.31 \\

+ RL: file F1
& 67.29 & 72.67 & 64.77
& 33.09 & 9.53
& \textbf{54.00} & 52.81 & \textbf{67.07} \\

+ RL: file $F_2$
& 55.47 & \textbf{75.80} & 47.02
& 28.70 & 8.85
& 48.34 & \textbf{58.89} & 49.90 \\

+ RL: file + class + function F1
& 65.22 & 72.20 & 61.99
& \textbf{35.31} & \textbf{13.80}
& 53.10 & 52.97 & 64.64 \\

+ RL: file F1 + turn penalty
& \textbf{67.34} & 71.60 & \textbf{65.29}
& 32.48 & 10.28
& 53.37 & 51.83 & 66.51 \\

\bottomrule
\end{tabular}
}

\caption{
    Effect of RL reward design on localization performance for Qwen3-1.7B.
    Results are reported on SWE-bench Lite and Multi-SWE-bench Flash.
}
\label{tab:rewards}
\vspace{-1em}
\end{table*}

\subsection{Reinforcement Learning Dynamics}
\label{app:rl-dynamics}

\begin{figure}[!htbp]
\centering
\includegraphics[width=\textwidth]{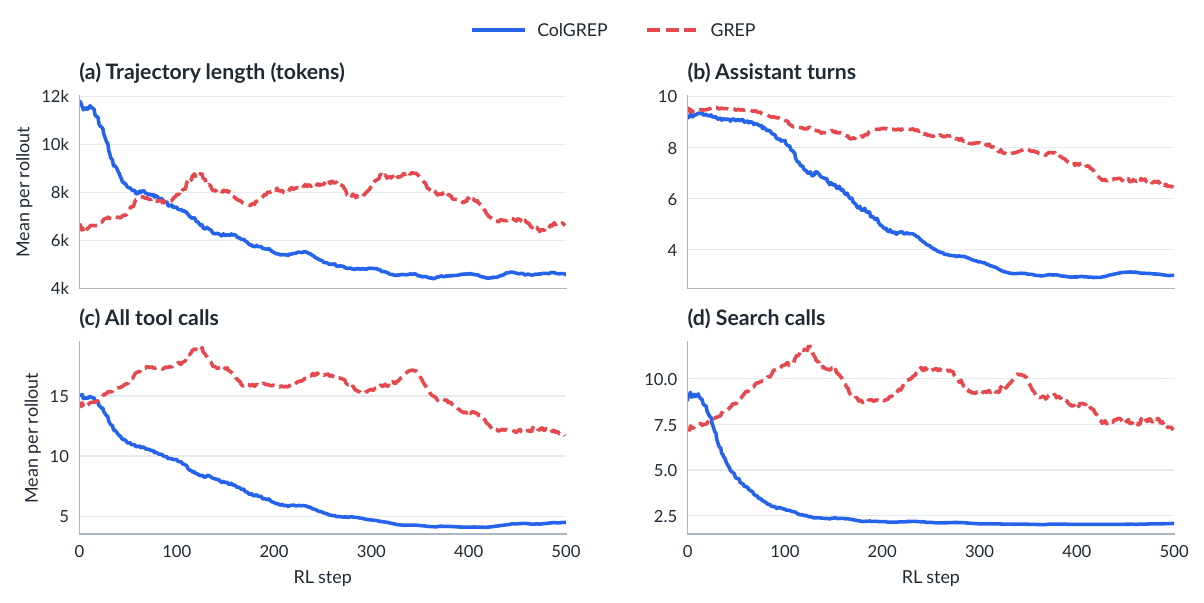}
\caption{
    \textbf{Evolution} of mean trajectory length, assistant turns, tool calls, and search calls per rollout for Qwen3-1.7B over 500 RL steps with a turn penalty, using the \grep and \colgrep harnesses.
}
\label{fig:rl-training-evolution}
\vspace{-1em}
\end{figure}

Figure~\ref{fig:rl-training-evolution} tracks the evolution of trajectory length, assistant turns, tool calls, and search calls over 500 RL optimization steps for Qwen3-1.7B when training with a turn penalty. We report the dynamics for both the \grep and \colgrep harnesses.

\subsection{Latency Measurement Protocol}
\label{app:latency-protocol}

Latency was measured on hosts with two Intel Xeon Platinum 8462Y+ processors, using 32 CPU threads for model inference. GPU inference used one NVIDIA H100 80,GB with vLLM 0.23.0 and BF16 weights; CPU inference used llama.cpp with BF16 GGUF weights. ColGREP ran on CPU with four threads in both settings. Each configuration processed 300 instances per benchmark sequentially; reported values average the two benchmark means equally. \grep and \colgrep used the same host within each benchmark and backend. Model weights, repositories, and precomputed indexes were preloaded into memory, with inference caching enabled. Timing covered the complete agent session, including search subprocess startup and retriever loading, but excluded staging, server startup, warm-up, and indexing.

\subsection{Evaluation Prompts}
\label{sec:harness_prompts}

\begingroup

\begin{table}[!htbp]
\centering
\small
\setlength{\tabcolsep}{5pt}
\renewcommand{\arraystretch}{1.08}
\arrayrulecolor{black!45}
\begin{tabular}{|p{0.94\columnwidth}|}
\hline
\cellcolor{black!8}
\begin{minipage}[c][2.6em][c]{0.90\columnwidth}
\centering\sffamily\bfseries
ColGREP system prompt
\end{minipage} \\
\hline
\cellcolor{white}
\begin{minipage}{0.90\columnwidth}
\vspace{0.6em}
\ttfamily\footnotesize\raggedright
You are a code-search agent. Given a software problem statement, find the exact code locations \textemdash{} files and line ranges \textemdash{} a developer must read and edit to solve it. This is a RETRIEVAL task: locate the relevant context; do NOT write a fix.\par\vspace{0.45em}
Follow this method:\par
\textbullet{} Start broad: use semantic search to survey the codebase and identify candidate files related to the problem.\par
\textbullet{} Narrow down: refine searches with path scopes, file-type filters, or regex pre-filters to isolate the most relevant code.\par
\textbullet{} Read and verify: open promising files and read the exact lines to confirm they are the right locations.\par
\textbullet{} Iterate: if results are too broad or off-target, reformulate your query from a different angle and search again.\par\vspace{0.45em}
You MUST search the repository before answering. Never answer from prior knowledge.\par\vspace{0.45em}
\#\#\# Finishing\par
When you have identified the relevant locations, you MUST use the \textasciigrave{}finish\textasciigrave{} tool to submit your answer. Do NOT simply reply with text \textemdash{} call the tool.\par
The \textasciigrave{}finish\textasciigrave{} tool expects a \textasciigrave{}locations\textasciigrave{} list; each entry is a string:\par
- \textasciigrave{}path/to/file\textasciigrave{}\par
- \textasciigrave{}path/to/file:START-END\textasciigrave{} (line range)\par
- \textasciigrave{}path/to/file:ClassName.method\_name\textasciigrave{} (symbol)\par
Call \textasciigrave{}finish\textasciigrave{} once when done. Do not keep searching after you know the answer.\par
\vspace{0.6em}
\end{minipage}
\\
\hline
\end{tabular}
\caption{Complete system prompt for the ColGREP code-localization configuration.}
\label{tab:eval_colgrep_prompt}
\end{table}

\begin{table}[!htbp]
\centering
\small
\setlength{\tabcolsep}{5pt}
\renewcommand{\arraystretch}{1.08}
\arrayrulecolor{black!45}
\begin{tabular}{|p{0.94\columnwidth}|}
\hline
\cellcolor{black!8}
\begin{minipage}[c][2.6em][c]{0.90\columnwidth}
\centering\sffamily\bfseries
Submission reminder
\end{minipage} \\
\hline
\cellcolor{white}
\begin{minipage}{0.90\columnwidth}
\vspace{0.6em}
\ttfamily\footnotesize\raggedright
You have 2 turns left. You MUST call the \textasciigrave{}finish\textasciigrave{} tool now with your best \textasciigrave{}locations\textasciigrave{} list \textemdash{} an unsubmitted session scores zero. Do not run another search.\par
\vspace{0.6em}
\end{minipage}
\\
\hline
\end{tabular}
\caption{User-role reminder with two turns remaining. The count and singular/plural form vary with the remaining budget.}
\label{tab:eval_submission_reminder}
\end{table}

\begin{table}[!htbp]
\centering
\small
\setlength{\tabcolsep}{5pt}
\renewcommand{\arraystretch}{1.08}
\arrayrulecolor{black!45}
\begin{tabular}{|p{0.94\columnwidth}|}
\hline
\cellcolor{black!8}
\begin{minipage}[c][2.6em][c]{0.90\columnwidth}
\centering\sffamily\bfseries
grep system prompt
\end{minipage} \\
\hline
\cellcolor{white}
\begin{minipage}{0.90\columnwidth}
\vspace{0.6em}
\ttfamily\footnotesize\raggedright
You are a code-search agent. Given a software problem statement, find the exact code locations \textemdash{} files and line ranges \textemdash{} a developer must read and edit to solve it. This is a RETRIEVAL task: locate the relevant context; do NOT write a fix.\par\vspace{0.45em}
Follow this method:\par
\textbullet{} Start from the task's concrete tokens: grep for identifiers, error strings, function or class names, and other literal text the problem mentions (e.g. \textasciigrave{}grep -rn "TokenError" .\textasciigrave{}).\par
\textbullet{} Narrow down: refine patterns with path scopes or file filters to isolate the most relevant code.\par
\textbullet{} Read and verify: open promising files and read the exact lines to confirm they are the right locations.\par
\textbullet{} Iterate: if results are too broad or off-target, reformulate your pattern from a different angle and search again.\par\vspace{0.45em}
You MUST search the repository before answering. Never answer from prior knowledge.\par\vspace{0.45em}
\#\#\# Finishing\par
When you have identified the relevant locations, you MUST use the \textasciigrave{}finish\textasciigrave{} tool to submit your answer. Do NOT simply reply with text \textemdash{} call the tool.\par
The \textasciigrave{}finish\textasciigrave{} tool expects a \textasciigrave{}locations\textasciigrave{} list; each entry is a string:\par
- \textasciigrave{}path/to/file\textasciigrave{}\par
- \textasciigrave{}path/to/file:START-END\textasciigrave{} (line range)\par
- \textasciigrave{}path/to/file:ClassName.method\_name\textasciigrave{} (symbol)\par
Call \textasciigrave{}finish\textasciigrave{} once when done. Do not keep searching after you know the answer.\par
\vspace{0.6em}
\end{minipage}
\\
\hline
\end{tabular}
\caption{Complete system prompt for the grep code-localization configuration.}
\label{tab:eval_grep_prompt}
\end{table}

\begin{table}[!htbp]
\centering
\small
\setlength{\tabcolsep}{5pt}
\renewcommand{\arraystretch}{1.08}
\arrayrulecolor{black!45}
\begin{tabular}{|p{0.94\columnwidth}|}
\hline
\cellcolor{black!8}
\begin{minipage}[c][2.6em][c]{0.90\columnwidth}
\centering\sffamily\bfseries
ColGREP tool descriptions
\end{minipage} \\
\hline
\cellcolor{white}
\begin{minipage}{0.90\columnwidth}
\vspace{0.6em}
\ttfamily\footnotesize\raggedright
\textbf{TOOL: colgrep}\par
Semantic code search over the repository (finds code by MEANING, not just text). Give a natural-language query describing the code this task needs.\par\vspace{0.25em}
\textbf{query} (string; optional): Natural-language description, e.g. "where session token expiry is validated".\par
\textbf{k} (integer; optional): Number of results to return (default 10, max 25).\par
\textbf{path} (string; optional): Optional subdirectory or file to scope the search.\par
\textbf{include} (string; optional): Optional file glob, e.g. "*.py".\par
\textbf{pattern} (string; optional): Optional grep-style regex pre-filter (colgrep -e), combined with \textasciigrave{}query\textasciigrave{}: the regex narrows the candidate documents, the query ranks them semantically. Use for exact names, numbers, or rare terms, e.g. "Art Deco|South Beach". Case-insensitive; if it matches nothing, retry without it.\par
\textbf{command} (string; optional): Alternative: a full colgrep CLI string.\par
\par\vspace{0.6em}
\textbf{TOOL: terminal}\par
Run one read-only shell command (ls/cd/cat/head/tail/pwd/grep/find/\ldots{}). Use colgrep for semantic search.\par\vspace{0.25em}
\textbf{command} (string; required): One shell command.\par
\par\vspace{0.6em}
\textbf{TOOL: finish}\par
Submit your final answer: the code locations a developer must read and edit. Call this once, when you are done searching. Each location is a string: 'path/to/file', 'path/to/file:START-END' (line range), or 'path/to/file:ClassName.method\_name' (symbol).\par\vspace{0.25em}
\textbf{locations} (array of strings; required): The relevant locations, most important first.\par
\vspace{0.6em}
\end{minipage}
\\
\hline
\end{tabular}
\caption{Complete tool descriptions and parameters exposed to the ColGREP agent. All \texttt{colgrep} arguments are optional in the schema; \texttt{command} provides the alternative CLI interface.}
\label{tab:eval_colgrep_tool}
\end{table}

\begin{table}[!htbp]
\centering
\small
\setlength{\tabcolsep}{5pt}
\renewcommand{\arraystretch}{1.08}
\arrayrulecolor{black!45}
\begin{tabular}{|p{0.94\columnwidth}|}
\hline
\cellcolor{black!8}
\begin{minipage}[c][2.6em][c]{0.90\columnwidth}
\centering\sffamily\bfseries
grep tool descriptions
\end{minipage} \\
\hline
\cellcolor{white}
\begin{minipage}{0.90\columnwidth}
\vspace{0.6em}
\ttfamily\footnotesize\raggedright
\textbf{TOOL: terminal}\par
Run one read-only shell command (grep/find/ls/cd/cat/head/tail/pwd/\ldots{}). Search the repository with grep, e.g. \textasciigrave{}grep -rn "TokenError" .\textasciigrave{}.\par\vspace{0.25em}
\textbf{command} (string; required): One shell command.\par
\par\vspace{0.6em}
\textbf{TOOL: finish}\par
Submit your final answer: the code locations a developer must read and edit. Call this once, when you are done searching. Each location is a string: 'path/to/file', 'path/to/file:START-END' (line range), or 'path/to/file:ClassName.method\_name' (symbol).\par\vspace{0.25em}
\textbf{locations} (array of strings; required): The relevant locations, most important first.\par
\vspace{0.6em}
\end{minipage}
\\
\hline
\end{tabular}
\caption{Complete tool descriptions and parameters exposed to the grep agent. Lexical searches are passed to \texttt{terminal}.}
\label{tab:eval_grep_tool}
\end{table}

\endgroup
\FloatBarrier
%%%%%%%%%%%%%%%%%%%%%%%%%%%%%%%%%%%%%%%%%%%%%%%

\subsection{Examples of Colgrep vs. Grep trajectories}
\label{sec:trajectory-examples}

\begingroup
\definecolor{trajBlue}{HTML}{245B82}
\definecolor{trajOchre}{HTML}{896025}
\definecolor{trajInk}{HTML}{263440}
\definecolor{trajGray}{HTML}{F3F5F7}
\tcbset{trajbubble/.style={reset,
  colback=white,colframe=black!15,boxrule=0.35pt,arc=2pt,
  left=5pt,right=5pt,top=2pt,bottom=2pt,
  before skip=1pt,after skip=1pt,
  fontupper=\fontencoding{T1}\fontfamily{ptm}\fontseries{m}\fontshape{n}\fontsize{8.5}{10}\selectfont\raggedright,
  fonttitle=\fontencoding{T1}\fontfamily{phv}\fontseries{b}\fontshape{n}\selectfont\fontsize{8}{9}\selectfont,
  coltitle=trajInk,colbacktitle=trajGray,toptitle=1pt,bottomtitle=1pt}}
\lstset{basicstyle=\fontencoding{T1}\fontfamily{pcr}\fontseries{m}\fontshape{n}\fontsize{7.5}{8.8}\selectfont,
  numbers=none,frame=none,backgroundcolor={},rulecolor=\color{black},
  xleftmargin=0pt,xrightmargin=0pt,resetmargins=true,
  showspaces=false,showtabs=false,tabsize=2,
  keywordstyle={},commentstyle={},stringstyle={},identifierstyle={},
  escapeinside={},escapechar={},texcl=false,literate={},
  breakindent=0pt,breakautoindent=true,prebreak={},postbreak={},
  columns=fullflexible,keepspaces=true,breaklines=true,
  breakatwhitespace=false,showstringspaces=false,
  aboveskip=2pt,belowskip=0pt,mathescape=false,
  language={},upquote=true}
\def\trajcut{\par{\color{black!60}\fontsize{7.5}{9}\selectfont [...]}\par}
\def\trajskip#1{%
  \par\smallskip{\centering\color{black!60}\fontsize{7.5}{9}\selectfont
  [#1 omitted]\par}\smallskip}

% Fixed design coordinates: the same line wrapping in every host document.
% All content is native LaTeX; no external PDF or image is needed.
\ifdefined\trajcomparisonbox\else\newsavebox{\trajcomparisonbox}\fi
\ifdefined\trajcomparisonfitbox\else\newsavebox{\trajcomparisonfitbox}\fi
\def\trajresetlayout{%
  \renewcommand{\baselinestretch}{1}%
  \renewcommand{\rmdefault}{ptm}%
  \renewcommand{\sfdefault}{phv}%
  \renewcommand{\ttdefault}{pcr}%
  \fontencoding{T1}\normalfont\fontsize{10}{12}\selectfont
  \color{black}%
  \setlength{\parindent}{0pt}%
  \setlength{\parskip}{0pt}%
  \setlength{\leftskip}{0pt}%
  \setlength{\rightskip}{0pt}%
  \setlength{\tabcolsep}{6pt}%
}
% Keep the caption inside the fitted composition; host caption styles do not
% change its height or spacing. The normal figure counter/label still work.
\makeatletter
\long\def\@makecaption#1#2{%
  \vskip 8pt
  {\fontencoding{T1}\fontfamily{ptm}\fontseries{m}\fontshape{n}%
   \fontsize{8.5}{10}\selectfont\noindent #1: #2\par}%
}
\makeatother
\def\trajshowcomposition{%
  \ifdim\linewidth<6.5in
    \sbox{\trajcomparisonfitbox}{\resizebox{\linewidth}{!}{\usebox{\trajcomparisonbox}}}%
  \else
    \sbox{\trajcomparisonfitbox}{\usebox{\trajcomparisonbox}}%
  \fi
  \ifdim\dimexpr\ht\trajcomparisonfitbox+\dp\trajcomparisonfitbox\relax>0.88\textheight
    \resizebox*{!}{0.88\textheight}{\usebox{\trajcomparisonfitbox}}%
  \else
    \usebox{\trajcomparisonfitbox}%
  \fi
}

\begin{figure*}[!htbp]
\centering
\begin{lrbox}{\trajcomparisonbox}%
\begin{minipage}{6.5in}
\trajresetlayout
\centering
% Source: trajectory_example_xarray_4248.md (paired seed-1 transcripts).
\begin{tcolorbox}[trajbubble,title={User (problem statement): pydata\_\_xarray-4248}]
Feature request: show units in dataset overview
\par\smallskip
Here's a hypothetical dataset:
\par
It would be really nice if the units of the coordinates and of the data variables were shown in the \texttt{Dataset} repr, for example as:
\par\smallskip
\begin{minipage}[t]{0.485\linewidth}
\textbf{Original display}
\begin{lstlisting}
<xarray.Dataset>
Dimensions:  (time: 3, x: 988, y: 822)
Coordinates:
  * x         (x) float64 ...
  * y         (y) float64 ...
  * time      (time) datetime64[ns] ...
Data variables:
    rainfall  (time, y, x) float32 ...
    max_temp  (time, y, x) float32 ...
\end{lstlisting}
\end{minipage}\hfill
\begin{minipage}[t]{0.485\linewidth}
\textbf{Requested display}
\begin{lstlisting}
<xarray.Dataset>
Dimensions:  (time: 3, x: 988, y: 822)
Coordinates:
  * x, in metres      (x) float64 ...
  * y, in metres      (y) float64 ...
  * time              (time) datetime64[ns] ...
Data variables:
    rainfall, in mm    (time, y, x) float32 ...
    max_temp, in deg C (time, y, x) float32 ...
\end{lstlisting}
\end{minipage}
\par\smallskip
Find the code locations relevant to this problem.
\end{tcolorbox}
\begin{minipage}[t]{0.485\textwidth}
\vspace{0pt}
\begin{tcolorbox}[trajbubble,colback=trajBlue!7,colframe=trajBlue!40,top=5pt,bottom=5pt]
{\sffamily\fontsize{10}{12}\selectfont\bfseries\color{trajBlue}colgrep harness}\hfill
{\sffamily\fontsize{8}{9}\selectfont 4 assistant turns}
\end{tcolorbox}
\begin{tcolorbox}[trajbubble,title={Agent (Turn 1)},colframe=trajBlue!35]
\begin{lstlisting}
{"query": "show units in dataset overview repr coordinates data variables"}
{"query": "dataset repr units dimensions coordinates variables"}
\end{lstlisting}
\end{tcolorbox}
\begin{tcolorbox}[trajbubble,title={Tool (Turn 1; excerpts)},colback=trajGray]
\trajcut
\begin{lstlisting}
[6] xarray/core/formatting.py:256-678  (function dataset_repr)
\end{lstlisting}
\trajcut
\end{tcolorbox}
\begin{tcolorbox}[trajbubble,title={Agent (Turn 2)},colframe=trajBlue!35]
Let me look at the \texttt{dataset\_repr} function in formatting.py, which is the core of what needs to be edited.
\begin{lstlisting}
sed -n '256,678p' xarray/core/formatting.py
\end{lstlisting}
\begin{lstlisting}
sed -n '187,200p' xarray/core/dataset.py
\end{lstlisting}
\end{tcolorbox}
\begin{tcolorbox}[trajbubble,title={Tool (Turn 2; excerpts)},colback=trajGray]
\trajcut
\begin{lstlisting}
def summarize_variable(
    name: Hashable, var, col_width: int, marker: str = " ", max_width: int = None
):
    """Summarize a variable in one line, e.g., for the Dataset.__repr__."""
\end{lstlisting}
\trajcut
\begin{lstlisting}
    first_col = pretty_print(f"  {marker} {name} ", col_width)
    if var.dims:
        dims_str = "({}) ".format(", ".join(map(str, var.dims)))
    else:
        dims_str = ""
    front_str = f"{first_col}{dims_str}{var.dtype} "
\end{lstlisting}
\trajcut
\end{tcolorbox}
\begin{tcolorbox}[trajbubble,title={Agent (Turn 3)},colframe=trajBlue!35]
Let me see the rest of the \texttt{dataset\_repr} function, particularly the part that builds the dataset summary.
\trajcut
\begin{lstlisting}
sed -n '678,720p' xarray/core/formatting.py
\end{lstlisting}
\end{tcolorbox}
\begin{tcolorbox}[trajbubble,title={Tool (Turn 3; excerpts)},colback=trajGray]
\begin{lstlisting}
return "\n".join(summary)
\end{lstlisting}
\end{tcolorbox}
\begin{tcolorbox}[trajbubble,title={Agent (Turn 4)},colframe=trajBlue!35]
I have all the relevant locations.
\trajcut
\par\smallskip{\sffamily\bfseries\fontsize{8}{9}\selectfont finish}
\begin{lstlisting}
{"locations": ["xarray/core/formatting.py:256-678"]}
\end{lstlisting}
\end{tcolorbox}
\begin{tcolorbox}[trajbubble,colback=trajBlue!7,colframe=trajBlue!35]
\textbf{File F1: 1.00} \enspace Exact match to the gold file set.
\end{tcolorbox}
\end{minipage}
\hfill
\begin{minipage}[t]{0.485\textwidth}
\vspace{0pt}
\begin{tcolorbox}[trajbubble,colback=trajOchre!7,colframe=trajOchre!40,top=5pt,bottom=5pt]
{\sffamily\fontsize{10}{12}\selectfont\bfseries\color{trajOchre}grep harness}\hfill
{\sffamily\fontsize{8}{9}\selectfont 7 assistant turns}
\end{tcolorbox}
\begin{tcolorbox}[trajbubble,title={Agent (Turn 1)},colframe=trajOchre!35]
\begin{lstlisting}
grep -rn "unit" xarray/ --include=*.py
\end{lstlisting}
\begin{lstlisting}
grep -rn "repr" xarray/ --include=*.py
\end{lstlisting}
\end{tcolorbox}
\begin{tcolorbox}[trajbubble,title={Tool (Turn 1; excerpts)},colback=trajGray]
\begin{lstlisting}
xarray/backends/api.py:386:        If True, decode variables and coordinates with time units in
\end{lstlisting}
\trajcut
\begin{lstlisting}
xarray/backends/api.py:382:        represented using ``np.datetime64[ns]`` objects.  If False, always
\end{lstlisting}
\trajcut
\end{tcolorbox}
\trajskip{Turns 2--4}
\begin{tcolorbox}[trajbubble,title={Agent (Turn 5)},colframe=trajOchre!35]
\begin{lstlisting}
grep -rn "units" xarray/core/
grep -rln "__repr__" xarray/core/
\end{lstlisting}
\end{tcolorbox}
\begin{tcolorbox}[trajbubble,title={Tool (Turn 5; excerpts)},colback=trajGray]
\trajcut
\begin{lstlisting}
xarray/core/dataset.py:4800:                                  'attrs': {'units':'s'}}},
\end{lstlisting}
\trajcut
\begin{lstlisting}
xarray/core/dataset.py
xarray/core/formatting.py
\end{lstlisting}
\trajcut
\end{tcolorbox}
\begin{tcolorbox}[trajbubble,title={Agent (Turn 6)},colframe=trajOchre!35]
I have a good understanding. The key locations are in \texttt{xarray/\allowbreak core/\allowbreak dataset.py} where the Dataset's \_\_repr\_\_ is defined, and where units might be extracted from variables.
\trajcut
\begin{lstlisting}
sed -n 4750,4820p xarray/core/dataset.py
\end{lstlisting}
\begin{lstlisting}
grep -rln "units" xarray/core/dataset.py
\end{lstlisting}
\end{tcolorbox}
\begin{tcolorbox}[trajbubble,title={Tool (Turn 6; excerpts)},colback=trajGray]
\trajcut
\begin{lstlisting}
def to_dict(self, data=True):
    """
    Convert this dataset to a dictionary following xarray naming
    conventions.
\end{lstlisting}
\trajcut
\begin{lstlisting}
xarray/core/dataset.py
\end{lstlisting}
\end{tcolorbox}
\begin{tcolorbox}[trajbubble,title={Agent (Turn 7)},colframe=trajOchre!35]
I have all the information needed.
\trajcut
\par\smallskip{\sffamily\bfseries\fontsize{8}{9}\selectfont finish}
\begin{lstlisting}
{"locations": ["xarray/core/dataset.py:4750-4820"]}
\end{lstlisting}
\end{tcolorbox}
\begin{tcolorbox}[trajbubble,colback=trajOchre!7,colframe=trajOchre!35]
\textbf{File F1: 0.00} \enspace Submitted file differs from gold.
\end{tcolorbox}
\end{minipage}
\small
\caption{\fontsize{8.5}{10}\selectfont \textbf{Example 1 --- Selected ColGREP vs. Grep trajectory excerpts.} Qwen3-1.7B trained with weighted SFT + RL (file F1 with turn penalty; step 500). ColGREP reaches the gold file, \texttt{xarray/core/formatting.py}; grep follows a units match into the \texttt{to\_dict} documentation in \texttt{dataset.py}. Excerpts retain the original turn numbers.}
\label{fig:trajectory-colgrep-grep}
\end{minipage}%
\end{lrbox}%
\trajshowcomposition
\end{figure*}

% A2: Go, cli__cli-2279.
% Source: trajectory_examples_qwen3_1.7b.md, section A2.
\begin{figure*}[!htbp]
\centering
\begin{lrbox}{\trajcomparisonbox}%
\begin{minipage}{6.5in}
\trajresetlayout
\centering
\begin{tcolorbox}[trajbubble,title={User (problem statement): cli\_\_cli-2279}]
add message on \texttt{pr checks} returns no CI with non-empty PR
\trajcut
Update messaging when \texttt{gh pr check} run on repo w/o CI

When running this \texttt{gh pr checks [branch] -R [owner/\allowbreak repo]} command and there is no pull request but no CI, I get this:

\texttt{no open pull requests found for branch "JJJ"}

However, when running the same command on a different branch and there is a pull request but no CI, nothing is returned. 

In both cases, I think it would be more accurate to say something like "no CI checks are set up for this repository".
\par\smallskip
Find the code locations relevant to this problem.
\end{tcolorbox}
\begin{minipage}[t]{0.485\textwidth}
\vspace{0pt}
\begin{tcolorbox}[trajbubble,colback=trajBlue!7,colframe=trajBlue!40,top=5pt,bottom=5pt]
{\sffamily\fontsize{10}{12}\selectfont\bfseries\color{trajBlue}colgrep harness}\hfill
{\sffamily\fontsize{8}{9}\selectfont 4 assistant turns}
\end{tcolorbox}
\begin{tcolorbox}[trajbubble,title={Agent (Turn 1)},colframe=trajBlue!35]
\begin{lstlisting}
{"query": "pr checks returns no CI with non-empty PR message"}
{"query": "gh pr checks command no CI checks setup"}
\end{lstlisting}
\end{tcolorbox}
\begin{tcolorbox}[trajbubble,title={Tool (Turn 1; excerpts)},colback=trajGray]
\trajcut
\begin{lstlisting}
[2] pkg/cmd/pr/checks/checks.go:1-227  (function NewCmdChecks)
\end{lstlisting}
\trajcut
\begin{lstlisting}
[1] pkg/cmd/pr/checks/checks.go:1-227  (function NewCmdChecks)
\end{lstlisting}
\trajcut
\end{tcolorbox}
\begin{tcolorbox}[trajbubble,title={Agent (Turn 2)},colframe=trajBlue!35]
Let me read the relevant part of \texttt{pkg/cmd/pr/checks/checks.go} to understand the full flow and the exact code that needs to be updated.
\begin{lstlisting}
sed -n '1,60p' pkg/cmd/pr/checks/checks.go
\end{lstlisting}
\end{tcolorbox}
\begin{tcolorbox}[trajbubble,title={Tool (Turn 2; excerpt)},colback=trajGray]
\trajcut
\begin{lstlisting}
cmd := &cobra.Command{
  Use:   "checks [<number> | <url> | <branch>]",
  Short: "Show CI status for a single pull request",
\end{lstlisting}
\trajcut
\begin{lstlisting}
return checksRun(opts)
\end{lstlisting}
\trajcut
\end{tcolorbox}
\begin{tcolorbox}[trajbubble,title={Agent (Turn 3)},colframe=trajBlue!35]
\trajcut
Let me verify the exact line range of \texttt{checksRun}.
\begin{lstlisting}
sed -n '60,110p' pkg/cmd/pr/checks/checks.go
\end{lstlisting}
\end{tcolorbox}
\begin{tcolorbox}[trajbubble,title={Tool (Turn 3; excerpts)},colback=trajGray]
\trajcut
\begin{lstlisting}
if len(pr.Commits.Nodes) == 0 {
  return nil
}

rollup := pr.Commits.Nodes[0].Commit.StatusCheckRollup.Contexts.Nodes
if len(rollup) == 0 {
  return nil
}
\end{lstlisting}
\trajcut
\end{tcolorbox}
\begin{tcolorbox}[trajbubble,title={Agent (Turn 4)},colframe=trajBlue!35]
\trajcut
Let me finalize.
\par\smallskip{\sffamily\bfseries\fontsize{8}{9}\selectfont finish}
\begin{lstlisting}
{"locations": ["pkg/cmd/pr/checks/checks.go:60-110"]}
\end{lstlisting}
\end{tcolorbox}
\begin{tcolorbox}[trajbubble,colback=trajBlue!7,colframe=trajBlue!35]
\textbf{File F1: 1.00} \enspace Exact match to the gold file set.
\end{tcolorbox}
\end{minipage}
\hfill
\begin{minipage}[t]{0.485\textwidth}
\vspace{0pt}
\begin{tcolorbox}[trajbubble,colback=trajOchre!7,colframe=trajOchre!40,top=5pt,bottom=5pt]
{\sffamily\fontsize{10}{12}\selectfont\bfseries\color{trajOchre}grep harness}\hfill
{\sffamily\fontsize{8}{9}\selectfont 6 assistant turns}
\end{tcolorbox}
\begin{tcolorbox}[trajbubble,title={Agent (Turn 1)},colframe=trajOchre!35]
\begin{lstlisting}
grep -rn "pr checks" . --include=*.go
\end{lstlisting}
\begin{lstlisting}
grep -rn "no open pull requests" .
\end{lstlisting}
\end{tcolorbox}
\begin{tcolorbox}[trajbubble,title={Tool (Turn 1; excerpts)},colback=trajGray]
\begin{lstlisting}
(grep: ran, no output)
\end{lstlisting}
\begin{lstlisting}
./pkg/cmd/pr/status/status.go:130:    shared.PrintMessage(out, "  You have no open pull requests")
\end{lstlisting}
\trajcut
\end{tcolorbox}
\trajskip{Turns 2--4}
\begin{tcolorbox}[trajbubble,title={Agent (Turn 5)},colframe=trajOchre!35]
Now I understand the problem. The key location is the message logic in \texttt{status.go}.
\trajcut
\begin{lstlisting}
sed -n 112,140p pkg/cmd/pr/status/status.go
grep -rn "no CI checks" .
\end{lstlisting}
\end{tcolorbox}
\begin{tcolorbox}[trajbubble,title={Tool (Turn 5; excerpts)},colback=trajGray]
\trajcut
\begin{lstlisting}
shared.PrintHeader(out, "Created by you")
if prPayload.ViewerCreated.TotalCount > 0 {
  printPrs(out, prPayload.ViewerCreated.TotalCount, prPayload.ViewerCreated.PullRequests...)
} else {
  shared.PrintMessage(out, "  You have no open pull requests")
}
\end{lstlisting}
\trajcut
\begin{lstlisting}
(grep: ran, no output)
\end{lstlisting}
\end{tcolorbox}
\begin{tcolorbox}[trajbubble,title={Agent (Turn 6)},colframe=trajOchre!35]
\trajcut
The relevant locations are the message logic in \texttt{else if currentPRHeadRef == ""} and the \texttt{shared.PrintMessage} calls. Let me finalize the locations.
\par\smallskip{\sffamily\bfseries\fontsize{8}{9}\selectfont finish}
\begin{lstlisting}
{"locations": ["pkg/cmd/pr/status/status.go:112-124", "pkg/cmd/pr/status/status.go:125-140"]}
\end{lstlisting}
\end{tcolorbox}
\begin{tcolorbox}[trajbubble,colback=trajOchre!7,colframe=trajOchre!35]
\textbf{File F1: 0.00} \enspace Submitted file differs from gold.
\end{tcolorbox}
\end{minipage}
\small
\caption{\fontsize{8.5}{10}\selectfont \textbf{Example 2 --- Selected ColGREP vs. Grep trajectory excerpts.} Qwen3-1.7B, weighted SFT + RL (file F1 with turn penalty, step 500). ColGREP reaches the gold file, \texttt{pkg/cmd/pr/checks/checks.go}, and reads the empty-check branches; grep follows a similar message in \texttt{pr/status/status.go}. Excerpts retain the original turn numbers.}

\label{fig:trajectory-colgrep-grep-go}
\end{minipage}%
\end{lrbox}%
\trajshowcomposition
\end{figure*}

% A3: TypeScript, vuejs__core-8538.
% Source: trajectory_examples_qwen3_1.7b.md, section A3.
\begin{figure*}[!htbp]
\centering
\begin{lrbox}{\trajcomparisonbox}%
\begin{minipage}{6.5in}
\trajresetlayout
\centering
\begin{tcolorbox}[trajbubble,title={User (problem statement): vuejs\_\_core-8538}]
fix(compiler-sfc): take into account imports that are used within dynamic arguments
\trajcut
(compiler-sfc): \texttt{\textless{}script setup\textgreater{}} - Import usage analyzer doesn't resolve dynamic arguments in non-inline mode
\trajcut
\textbf{Steps to reproduce}
\par
1. Import an object/enum from another file.
\par
2. Try to use their keys as a source of dynamic attribute name or slot name in \texttt{\textless{}script setup\textgreater{}} component. Note that this enum/object is \textbf{not} supposed to be used anywhere else in the component except the mentioned place.
\par
3. Observe an error since our object/enum is not exposed to the template (you can also check the generated code and take a look at the final \texttt{\_\_returned\_\_} variable).
\par\smallskip
\textbf{What is expected?}\par
This object/enum should be properly exposed to the template in the final \texttt{\_\_returned\_\_} variable since it's already being used in the template.
\par\smallskip
\textbf{What is actually happening?}\par
An error like \texttt{Cannot read properties of undefined (reading 'someProperty')} because the variable we're trying to get access to is not exposed.
\trajcut
Find the code locations relevant to this problem.
\end{tcolorbox}
\begin{minipage}[t]{0.485\textwidth}
\vspace{0pt}
\begin{tcolorbox}[trajbubble,colback=trajBlue!7,colframe=trajBlue!40,top=5pt,bottom=5pt]
{\sffamily\fontsize{10}{12}\selectfont\bfseries\color{trajBlue}colgrep harness}\hfill
{\sffamily\fontsize{8}{9}\selectfont 3 assistant turns}
\end{tcolorbox}
\begin{tcolorbox}[trajbubble,title={Agent (Turn 1)},colframe=trajBlue!35]
\begin{lstlisting}
{"query": "compiler-sfc import usage analyzer resolve dynamic arguments non-inline mode"}
{"query": "script setup import usage dynamic argument resolution"}
\end{lstlisting}
\end{tcolorbox}
\begin{tcolorbox}[trajbubble,title={Tool (Turn 1; excerpts)},colback=trajGray]
\begin{lstlisting}
[1] packages/compiler-sfc/src/script/importUsageCheck.ts:1-115  (function isImportUsed)
\end{lstlisting}
\trajcut
\begin{lstlisting}
export function isImportUsed(local: string, sfc: SFCDescriptor): boolean {
\end{lstlisting}
\trajcut
\end{tcolorbox}
\begin{tcolorbox}[trajbubble,title={Agent (Turn 2)},colframe=trajBlue!35]
Let me read the key files:
\trajcut
\begin{lstlisting}
sed -n '1,80p' packages/compiler-sfc/src/compileScript.ts
\end{lstlisting}
\begin{lstlisting}
sed -n '1,120p' packages/compiler-sfc/src/script/importUsageCheck.ts
\end{lstlisting}
\end{tcolorbox}
\begin{tcolorbox}[trajbubble,title={Tool (Turn 2; excerpts)},colback=trajGray]
\trajcut
\begin{lstlisting}
for (let i = 0; i < node.props.length; i++) {
  const prop = node.props[i]
  if (prop.type === NodeTypes.DIRECTIVE) {
\end{lstlisting}
\trajcut
\begin{lstlisting}
if (prop.exp) {
  code += `,${processExp(
    (prop.exp as SimpleExpressionNode).content,
    prop.name
  )}`
}
\end{lstlisting}
\trajcut
\end{tcolorbox}
\begin{tcolorbox}[trajbubble,title={Agent (Turn 3)},colframe=trajBlue!35]
I have identified the key locations.
\trajcut
\par\smallskip{\sffamily\bfseries\fontsize{8}{9}\selectfont finish}
\begin{lstlisting}
{"locations": ["packages/compiler-sfc/src/script/importUsageCheck.ts:1-115"]}
\end{lstlisting}
\end{tcolorbox}
\begin{tcolorbox}[trajbubble,colback=trajBlue!7,colframe=trajBlue!35]
\textbf{File F1: 1.00} \enspace Exact match to the gold file set.
\end{tcolorbox}
\end{minipage}
\hfill
\begin{minipage}[t]{0.485\textwidth}
\vspace{0pt}
\begin{tcolorbox}[trajbubble,colback=trajOchre!7,colframe=trajOchre!40,top=5pt,bottom=5pt]
{\sffamily\fontsize{10}{12}\selectfont\bfseries\color{trajOchre}grep harness}\hfill
{\sffamily\fontsize{8}{9}\selectfont 8 assistant turns}
\end{tcolorbox}
\begin{tcolorbox}[trajbubble,title={Agent (Turn 1)},colframe=trajOchre!35]
\begin{lstlisting}
grep -rn "dynamic arguments" .
\end{lstlisting}
\begin{lstlisting}
grep -rn "import usage analyzer" .
\end{lstlisting}
\end{tcolorbox}
\begin{tcolorbox}[trajbubble,title={Tool (Turn 1; excerpts)},colback=trajGray]
\begin{lstlisting}
(grep: ran, no output)
\end{lstlisting}
\begin{lstlisting}
(grep: ran, no output)
\end{lstlisting}
\end{tcolorbox}
\trajskip{Turns 2--3}
\begin{tcolorbox}[trajbubble,title={Agent (Turn 4)},colframe=trajOchre!35]
The bug is about dynamic arguments in \texttt{\textless{}script setup\textgreater{}} where imports are used in non-inline mode.
\trajcut
\begin{lstlisting}
grep -rn "dynamic arguments" packages/compiler-sfc/src/
\end{lstlisting}
\begin{lstlisting}
grep -rn "__returned__" packages/compiler-sfc/src/
\end{lstlisting}
\end{tcolorbox}
\begin{tcolorbox}[trajbubble,title={Tool (Turn 4; excerpts)},colback=trajGray]
\begin{lstlisting}
(grep: ran, no output)
\end{lstlisting}
\begin{lstlisting}
packages/compiler-sfc/src/compileScript.ts:926:      `\nconst __returned__ = ${returned}\n` +
\end{lstlisting}
\trajcut
\end{tcolorbox}
\trajskip{Turns 5--7}
\begin{tcolorbox}[trajbubble,title={User (harness reminder after Turn 7)}]
You have 2 turns left. You MUST call the \texttt{finish} tool now with your best \texttt{locations} list --- an unsubmitted session scores zero. Do not run another search.
\end{tcolorbox}
\begin{tcolorbox}[trajbubble,title={Agent (Turn 8)},colframe=trajOchre!35]
\par\smallskip{\sffamily\bfseries\fontsize{8}{9}\selectfont finish}
\begin{lstlisting}
{"locations": ["packages/compiler-sfc/src/compileScript.ts:376-400", "packages/compiler-sfc/src/compileScript.ts:926-929"]}
\end{lstlisting}
\end{tcolorbox}
\begin{tcolorbox}[trajbubble,colback=trajOchre!7,colframe=trajOchre!35]
\textbf{File F1: 0.00} \enspace Submitted file differs from gold.
\end{tcolorbox}
\end{minipage}
\small
\caption{\fontsize{8.5}{10}\selectfont \textbf{Example 3 --- Selected ColGREP vs. Grep trajectory excerpts.} Qwen3-1.7B, weighted SFT + RL (file F1 with turn penalty, step 500). ColGREP reaches the gold file, \texttt{packages/compiler-sfc/src/script/importUsageCheck.ts}, and reads the directive traversal; grep submits \texttt{compileScript.ts} after repeated searches and a budget reminder. Excerpts retain the original turn numbers.}

\label{fig:trajectory-colgrep-grep-typescript}
\end{minipage}%
\end{lrbox}%
\trajshowcomposition
\end{figure*}
\endgroup